# Statistical properties and privacy guarantees of an original distance-based fully synthetic data generation method

Rémy Chapelle[1,2], Bruno Falissard[1]

[1]Centre de Recherche en Épidémiologie et Santé des Populations, Université Paris-Saclay, 16 avenue Paul Vaillant Couturier, Villejuif, 94800, France.
[2]École du Val-de-Grâce, 1 place Alphonse Laveran, Paris, 75005, France.

## Abstract

**Introduction:** The amount of data generated by original research is growing exponentially. Publicly releasing them is recommended to comply with the Open Science principles. However, data collected from human participants cannot be released as-is without raising privacy concerns. Fully synthetic data represent a promising answer to this challenge. This approach is explored by the French Centre de Recherche en Épidémiologie et Santé des Populations in the form of a synthetic data generation framework based on Classification and Regression Trees and an original distance-based filtering. The goal of this work was to develop a refined version of this framework and to assess its risk-utility profile with empirical and formal tools, including novel ones developed for the purpose of this evaluation.

**Materials and Methods:** Our synthesis framework consists of four successive steps, each of which is designed to prevent specific risks of disclosure. We assessed its performance by applying two or more of these steps to a rich epidemiological dataset. Privacy and utility metrics were computed for each of the resulting synthetic datasets, which were further assessed using machine learning approaches.

**Results:** Computed metrics showed a satisfactory level of protection against attribute disclosure attacks for each synthetic dataset, especially when the full framework was used. Membership disclosure attacks were formally prevented without significantly altering the data. Machine learning approaches showed a low risk of success for simulated singling out and linkability attacks. Distributional and inferential similarity with the original data were high with all datasets.

**Discussion:** This work showed the technical feasibility of generating publicly releasable synthetic data using a multi-step framework. Formal and empirical tools specifically developed for this demonstration are a valuable contribution to this field. Further research should focus on the extension and validation of these tools, in an effort to specify the intrinsic qualities of alternative data synthesis methods.

**Conclusion:** By successfully assessing the quality of data produced using a novel multi-step synthetic data generation framework, we showed the technical and conceptual soundness of the Open-CESP initiative, which seems ripe for full-scale implementation.

# Introduction

The ClinicalTrials.gov database is the largest clinical trial registry worldwide, and contained less than 2,000 trials at its creation in 2000. Twenty years later, this number has been multiplied by more than 200 (1). Annual publication statistics show similar trends in the epidemiological field (2). Most of the corresponding studies presumably generated original clinical and epidemiological data, which illustrates a massive expansion of research data in the healthcare domain.

The value of such data beyond original research projects has been increasingly recognized over the last decades, leading key authors and institutions to recommend their systematic sharing (3). The expected benefits of such initiatives include maximizing the value of collected data by encouraging multiple examinations and interpretations as well as minimizing duplicative efforts, hence reducing research costs and lowering human participants burden (4). Such considerations are of course not limited to health data, and various calls to share scientific data, results and frameworks are now part of a global "Open Science" movement (5,6).

Successful projects have already been developed under the Open Science banner. For example, the Global Human Settlement Layer project initiated by the European Joint Research Center has provided hundreds of researchers with open data to assess the sustainability of planet Earth (7). In the healthcare field, however, such achievements seem slower to materialize. This probably stems from privacy concerns, as health data are particularly prone to containing sensitive information (8).

In the last decades, several proposals have been made to address these concerns, mainly oriented around de-identifying health information (9), which in the broad sense means making published data impossible to link with a particular individual (10). In the statistical community, de-identification is also known as Statistical Disclosure Control (SDC) (11,12). In what follows, we will use the terms de-identification, anonymization and SDC indifferently, although other authors may give them distinct operational definitions.

Aggregating data to produce summary statistics is a straightforward method of de-identification. However, such data have less utility for research purposes than data at the individual level, also known as microdata (13). In essence, the publication of microdata is more aligned with the Open Science principles than the publication of aggregated data. Their de-identification is, however, tedious.

Historically, microdata de-identification has been performed through "anonymization operations". In these approaches, one or several privacy models are chosen, and the original data are altered to conform to them through successive row and column operations, including (if needed) noise addition. Possible privacy models include the so-called *k-anonymity* and *l-diversity*. Using them generally requires determining which variables of the microdata can be used to identify members of the underlying population. Depending on their identification capability, such variables are called explicit identifiers (e.g. social security number) or quasi-identifiers (e.g. date of birth). Additionally, some privacy models require users to identify sensitive variables, which in the case of health data could for example include disease or disability status (14). The core idea behind anonymization operations is then to remove (in part or totality) the information contained in identifiers and/or sensitive variables (14).

Despite their intuitive appeal, such methods have been subject to heavy criticisms in the past decades, not only because they tend to distort relationships between variables (15), but also because of their intrinsic vulnerability to various attacks (16). In brief, it can be shown that any adversary with enough background knowledge (possibly acquired from other independent databases) can re-identify subjects in data protected by such anonymization operations. That a potential adversary would care about collecting data from several databases to use them in combination is not a fictional idea, as illustrated by the 2008 Netflix Prize privacy breach (17). More recently, Roger et al. showed that 15 demographic variables were enough to identify 99.98% of Americans in any pseudonymized dataset (18). Such results certainly blur the line between identifiers and non-identifiers.

Because of these vulnerabilities and their resulting inability to ensure high levels of privacy, these models have been called syntactic privacy models, as opposed to semantic models which aim at providing formal privacy guarantees independently from any attacker's background knowledge (19). Semantic privacy models have been developed more recently than their syntactic counterparts, and essentially consist in variations around the concept of differential privacy developed by Dwork *et al.* in 2006 (19–21). In short, given a set of possible datasets $E$ and any set $F$, a randomized function $f: E \to F$ is said to be differentially private at an $\epsilon \in \mathbb{R}^{*+}$ level if, and only if (21):

$$\forall (S, D_1, D_2) \in \wp(K) \times E^2, \overline{D_1 \cap D_2} = \overline{D_1} - 1 = \overline{D_2} - 1, P(f(D_1) \in S) \leq \exp(\epsilon) \times P(f(D_2) \in S)$$

Intuitively, this means that a randomized function is differentially private at an $\epsilon$ level if modifying its input dataset by one element does not alter significantly its result. The desired level of significance is obtained by adjusting the value of $\epsilon$.

Differential privacy and its variant are often considered to offer the highest standards for data privacy (19,22), but are not flawless. Among others, differential privacy does not allow for direct publication of microdata (the randomized function in the formal description typically represents a mechanism generating aggregated data), is not directly compatible with standard statistical tools (attaining differential privacy requires noise addition that needs to be accounted for by statisticians) and requires users to choose a specific value for $\epsilon$, which is always arbitrary to some extent (23–29).

The lack of formal guarantees presented by classical privacy models and the impracticability of differential privacy for microdata has long suggested that a paradigmatic shift may be necessary to generate publicly releasable microdata (30). Synthetic data generation, which has gained attention in recent years, arguably represents such a paradigmatic shift.

Synthetic data can be defined as fictive data created from an actual dataset by a statistical process and designed for public consumption (31). Typically, a synthetized dataset resembles a real one, but contains no unit of the original data (cf. Figure 1). Alternatively, the synthesis process can be applied only to parts of the original dataset (e.g. some cells only). Such cases correspond to partially synthetic data, while the typical implementation described above is called fully synthetic data (32). In the following, we will use the terms "synthetic data" without explicit precision to refer to fully synthetic data only.

Original Data

| Gender | Height |
|--------|--------|
| F | 169 |
| M | 184 |
| F | 167 |
| F | 164 |
| M | 187 |
| F | 174 |

Synthetic Data

| Gender | Height |
|--------|--------|
| M | 191 |
| F | 156 |
| F | 176 |
| F | 168 |
| M | 185 |
| M | 183 |

*Figure 1. Fully synthetic data are created from original data and retain their statistical properties, but do not contain any of their rows.*

Donald Rubin and Roderick Little are generally credited for the first suggestion of using synthetic data as a SDC method (32). In their 1993 seminal papers (15,33), they proposed using the multiple imputation framework that they developed to handle missing data as a way of generating fictive microdata for general public use. Their idea was to preserve both the confidentiality of individuals from which the data originate (no synthetized unit being an actual unit) and the utility of the data (when properly configured, multiply imputed data can lead to valid statistical inferences). The nature of the data produced makes their confidentiality properties less reliant on syntactic privacy models (32) and may reduce the need for formal privacy guarantees such as those provided by differential privacy.

Thirty years later, these core principle still hold, but their application in the healthcare field is subject to several challenges. Many of these challenges are related to the determination of the best tradeoff between utility and disclosure risks of such data (34). When processing a dataset with SDC methods, there is indeed an inverse relationship between the confidentiality and the utility (practical value for general or specific purposes) of the resulting data (34,35). Synthetic data make no exception, with original data (highest utility but lowest confidentiality) and completely random data (highest confidentiality but null utility) lying at each extremity of the risk-utility (RU) curve.

In recent years, several metrics have been developed to help statisticians and computer scientists quantify the utility and disclosure risks of their synthetic data (34). They provide valuable information in the process of choosing the best RU profile for a synthetic data generation method, which still remains a NP-hard task (25). Other challenges related to the use of synthetic data include the determination of maximum acceptable levels of disclosure, which depends not only on statistical but also on methodological and social considerations (36–38).

Thirty years after Rubin's proposal, the maturation of this field has already allowed the publication of several synthetic datasets, such as synthetic versions of the French National Health Data System (SNDS) (39), the NIH National COVID Cohort Collaborative (40), and cancer data provided by the National Disease Registration Service (41). However, there currently seems to be no large-scale or institutional initiative to publish synthetic datasets generated from research data. In 2022, the French Centre de recherche en Epidémiologie et Santé des Populations (CESP) began to fill this gap by launching the Open-CESP project, with the aim of providing public access to synthetic datasets derived from research works led by the CESP. Four datasets have already been published as proofs of concept using a novel full synthesis framework based on Classification And Regression Trees (CART) and an original distance-based filtering.

In this context, the goal of our work was to develop a refined version of this framework, to discuss the content of the corresponding synthesizer and to assess its privacy and utility properties using various empirical and formal tools, including novel ones that we developed for the sake of this evaluation.

## Materials and Methods

Synthetic data generation methods primarily aim at preserving the utility of original data and the privacy of subjects from which they originate (42). To show how our framework perform on these two aspects, we computed privacy and utility metrics on the real and several synthetic versions of a rich dataset. We completed our assessment by performing standard utility evaluations on real-world analyses and using validated third-party libraries to quantify privacy risks. In the following subsections, we successively present our data synthesis framework, the original dataset that we used for its evaluation and the privacy and utility assessments conducted on synthetic versions of this dataset.

**Synthetic data generation framework**

The main steps of our Synthetic Data Generation Framework (SDGF) are depicted in Figure 2.

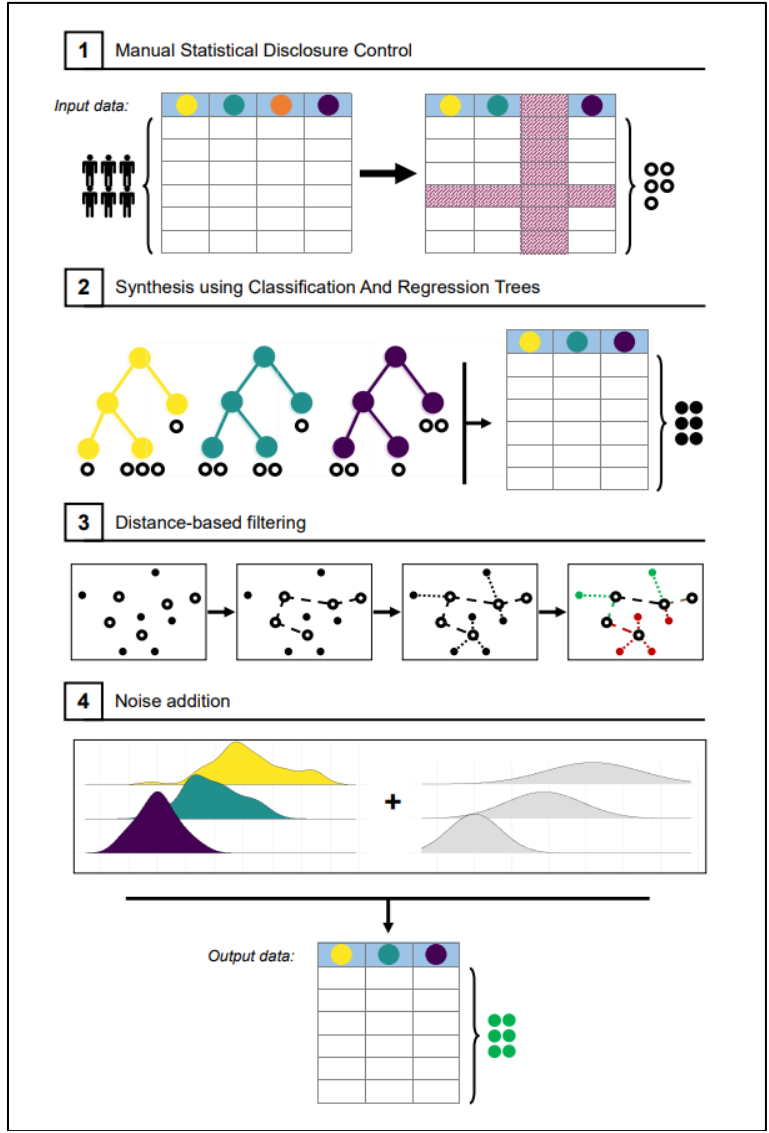

*Figure 2. Illustrated steps of our SDGF.*

Input data to our framework consist in microdata, with no particular restriction on the number of variables and observations besides those related to synthesis using CART (cf. infra). Input variables can be of any type, namely categorical or quantitative, and can be linked by any logical relationship. Longitudinal data are also supported. Successive steps of the framework are detailed in the following subsections. These steps comply with common guidelines on synthetic data generation (43), and even extend them on decisive points such as protection against some forms of data disclosure.

## 1. Manual SDC

Every input data treated by the Open-CESP team undergo manual SDC. In this first phase, classical anonymization operations may be applied, the nature and order of which depends on the exact type of data. After standard data management operations, data scientists and researchers typically sort variables into direct identifiers (linked to exactly one individual, for example social security numbers), quasi-identifiers (linked to a small number of individuals, for example town of residence) and sensitive outcomes (highly disclosive information, such as religious beliefs, that needs to be protected from potential intruders). Any variable which seems to have no scientific interest (which was not necessarily known when data was collected) is removed. Identifiers are also typically removed, even when they do not fall into this category. Other variables may be subject to categorization, bottom- or top-coding (for quantitative variables) and coarsening (for categorical variables), among others. Disclosure risk may be higher for statistical outliers and those may dropped (or their data perturbed) if necessary. The entire process is guided by scenarios of attack, where plausible behaviors of attackers trying to access sensitive outcomes are examined. For a detailed review on classical anonymization operations and related semantics, see e.g. (44) and (45).

When the addition of manual SDC and data synthesis seems insufficient to reach the desired level of privacy, no data are published. On the opposite, some input data may not need any manual SDC before being submitted to our synthesizer. Already published data, for example in peer-reviewed journals, are always taken into account in this manual SDC process.

## 2. Data synthesis

The core of every synthetic data generation method is the generative model used to produce data, also called the synthesizer. Multiple models have been reported in the literature, broadly classified into knowledge-driven and data-driven models (46). In a knowledge-driven approach, a model is manually created based on what is known of a particular population in terms of distributional and relational properties of target variables. This knowledge takes the form of generative rules, which can for example be implemented as multiple regression models and serve as a basis for the resulting synthesizer to generate data. Because of the fine tune needed to ensure the validity of these rules, creating a knowledge-driven model is a tedious work, generally implying preliminary literature reviews as well as the assistance of field researchers working on the subject involved. The main reward of this work is the ability of the resulting model to produce synthetic data without relying on real input data. This entails a minimal risk of information disclosure. However, this level of privacy comes at a substantial cost that we see as prohibitive: exclusively data-driven models are prone to miss a large part of the relationship between variables, simply because they have not been explicitly included in them (34). By choosing a knowledge-driven model, we would therefore deprive ourselves of one of the main benefits of synthetic data generation, that is their potential use to formulate new research hypotheses (47). Moreover, these methods can hardly be automated and therefore tend to be laborious and prone to human errors, which make them ill-suited to the publication of large amounts of data.

For all these reasons, we chose a data-driven synthesizer as part of our data generation method. This choice only slightly narrows the number of candidate models for our framework, as many different data-driven methods have been proposed in recent years (32). Unfortunately, there are currently no consensual guidelines on how to choose a data-driven generative model, nor formal ways of assessing the superiority of a given model against another one. In this context, we think that the a priori choice of a model for a framework should mainly depend on the technical constraints entailed by its intended usage. In the case of the Open-CESP project, aiming at publishing various datasets for general use of scientists (not necessarily familiar with synthetic data), the selected model should arguably:

1. Be probabilistic rather than deterministic, so that the synthesis algorithm can be published along datasets without raising privacy concerns.
2. Be suited to longitudinal as well as cross-sectional microdata, possibly including large numbers of variables.
3. Support logical constraints between variables to avoid anomalous values from being produced.
4. Be non-parametric in the classical sense of the term, which means that it must not be manually shaped to individual joint distributions in the original dataset.
5. Allow for coarse tuning so that for every input dataset, the best trade-off between utility of generated data and privacy of original units can be selected.
6. Be as easy to configure and computationally inexpensive as possible.

We think that these criteria make generative models based on CART the most a priori well-suited to our project. Most authors suggest that the appropriateness to use a particular generative model should also be assessed using specific metrics of utility and confidentiality (34,48,49). Models based on CART have been shown to perform well in this regard (36,50–53), and we confirmed these findings in previous analyses where we compared these models with other candidates on four testing datasets. These results strengthened our decision to use a synthesizer based on CART in our framework.

CART have been formally introduced by Leo Breiman in 1984 (54) as a tool to address decision problems. By essence, CART are binary trees in which every node contains observations of an original dataset meeting specific criteria related to some variables called predictors (55). These criteria are chosen such that the leaves of the tree are relatively homogeneous with respect to a given outcome variable, where the homogeneity is canonically understood in terms of a low Gini's index. This is done by machine learning implementations sometimes offering parameters to control for the number and location of the nodes. Resulting CART can then be used to determine the distribution of outcome variables conditionally on predictors, with a precision depending on these parameters.

Figure 3 shows the example of a CART generated from the dataset displayed in Figure 1.

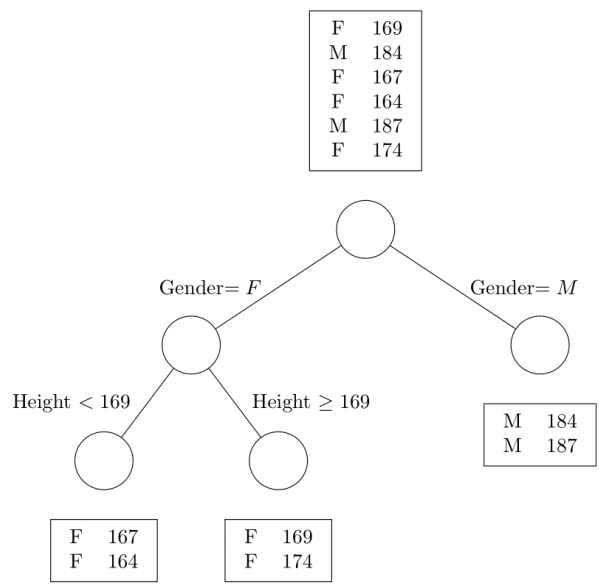

*Figure 3. Credible CART generated from data shown in Figure 1.*

In 2005, Jerome Reiter was the first to propose using CART to generate synthetic data (42). Although his initial work dealt with partial synthesis, it has then been extended to full synthesis (56,57). The core idea is to generate CART (which we refer to as $T_1, T_2, ..., T_n$) for each of the $n$ variables in the original dataset to capture their conditional distributions. Each original observation then undergoes the following synthesis process. First, a synthetic value for the variable $X_1$ is sampled among the values belonging to the same leaf of $T_1$. The location of the observation in $T_2, T_3, ..., T_n$) is then updated according to this new value, and the whole process is repeated for every variable in the dataset. In the end, it is very unlikely that a synthetic observation be identical to a real one. The probability of such a phenomenon is reduced as the number of observations in each leaf increases ; but doing so also decrease the precision of the conditional distribution estimates by CART and can lead to lower data utility. Different implementations of the algorithm can use other ways of reducing the resemblance between original and synthetic data, by example by setting a threshold on the intra-leaves variance. Several solutions can also be proposed to generate any number of synthetic units from an original dataset of limited length, such as preliminary samplings from the original data. General considerations on these points can be found in (42) and (57). In (42), Reiter also discusses the best choice for the order of variables in the synthesis process, as well as the sampling technique to be used within each leaf.

To generate synthetic data using CART, we use the *synthpop* R package (57), which is arguably one the most popular packages in this field. The empirical validity of datasets synthesized using *synthpop* for various use cases has been established by numerous studies (51,58–62), and its feasibility successfully assessed for longitudinal data (36). The package offers several parameters to tune the synthesis process, including the desired minimum number of units in any leaf of the CART, logical relationships between and desired number of synthetic units to be produced. Missing values are supported by considering them as a distinct modality of qualitative variables, and by the adjunction of an auxiliary variable for quantitative ones. Under the hood, *synthpop* can use distinct packages to generate CART, namely *rpart* (63) and *party* (64). Our preference goes to *rpart* because of its close alignment with Breiman's original ideas, conferring it a higher face validity in our views.

Generally, we generate synthetic datasets of same length than original ones, namely to prevent statistical conditions from being falsely met when performing parametric tests. This also helps controlling disclosure risk and artificial statistical power inflation. As for the number of observations in each leaf of generated CART, we see it as a major way of mitigating disclosure risk in our framework, as larger numbers of observations are associated with decreased resemblance between original and synthetic data. Consequently, the choice we make for this parameter highly depends on the results of previous and next steps of our framework, as well as the sensitivity of the input data.

Some authors recommend releasing multiple synthetic datasets for every original one (65). This idea mainly stems from the original proposal of Rubin for synthetic data generation (33), deeply rooted in its multiple imputation framework. The underlying principle is that no valid inference can be made from synthetic datasets if one does not take into account the additional variance produced by the synthesis process (66). Although the statistical reasoning seems solid, we think that this does not apply to the Open-CESP initiative, and consequently release only one synthetic version of each original dataset that we process. Our approach is indeed not meant to allow researchers to draw valid inferences from released datasets. Instead, we expect them to run analyses on synthetic data in an exploratory phase, and to confirm promising results on real data (available on demand). In our view, this last step is mandatory for the obtained results to be considered valid, and we strongly discourage presenting results obtained from synthetic datasets alone as conclusive. By doing so, we align with the policy of other data providers, such as the United States Census Bureau® which provides access to the SynLBD synthetic database but recommends running final analyses on original data (67). This policy prevents the inflation of disclosure risk that can arise if several datasets synthetized from the same original data are released incautiously. It also has the advantage of not requiring data users to use complicated statistical procedures to handle multiple datasets.

### 3. Distance-based filtering

Although terminologies vary between authors, attacks against anonymized microdata are generally categorized into membership disclosure attacks, identity disclosure attacks and attribute disclosure attacks (68,69). Table 1 outlines the main principles underlying each of these types of attacks.

| Type of attacks | Principle |
| --- | --- |
| Membership disclosure attack | Given the anonymized dataset and a sufficient amount of background knowledge, the attacker tries to establish that a given individual was part of the original dataset. |
| Identity disclosure attack | Given the anonymized dataset and a sufficient amount of background knowledge (often including membership knowledge), the attacker tries to identify a record in the dataset matching a given individual. |
| Attribute disclosure attack | Given the anonymized dataset and a sufficient amount of background knowledge (often including linkage knowledge), the attacker tries to infer unknown values of attributes about a given individual. Some authors reserve this term to inferences with a total certainty, and call "inferential disclosures" the inferences associated with a high, but less than 1, confidence (70). |

*Table 1. Characteristics of the main type of attacks against anonymized microdata.*

In our framework, formal protection against membership disclosure attacks is provided by noise addition performed in its last step and described in a subsequent section. On the other hand, manual SDC and the synthesis itself provide a significant level of protection against identification and attribute disclosure risks. In particular, fully synthetic microdata are often said to be immune from identification disclosure risk because there exists no meaningful link between rows of the original dataset and rows of the synthetic one (71). Although this point is valid at a theoretical level, CART-based synthesizers (like many other synthesizers) may generate rows closely resembling those of the original dataset (72). This phenomenon would arguably be of little interest to a potential attacker, for whom every synthetized row is the result of the same probabilistic mechanism and has the same negligible probability to belong to the original dataset. However, it is acknowledged that the presence of such rows in synthetic datasets can impair the trust of participants into the synthesis process (73). This is why we developed an algorithm capable of filtering out these rows based on their distance with other statistical units of the original and synthetic data. By doing so, we also aim at mitigating the risk of attribute disclosure, thereby addressing one of the main limitations of CART synthesizers which have been shown to provide high data utility, but also higher disclosure risks than alternative methods when used in isolation (74).

Our filtering method is based on four successive steps which are depicted in Figure 4.

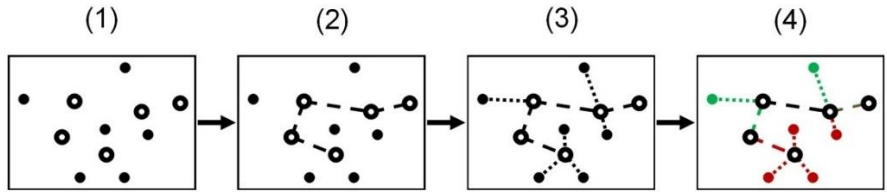

*Figure 4. Successive steps of the distance-based filtering algorithm. Empty circles represent original units, while solid circles represent synthetic ones. In step (4), green synthetic units are retained, while red ones are removed.*

Input data to the algorithm correspond to synthetic statistical units generated by CART as well as original units. As a first step, respective distances between all pairs of units are computed. To do so, we use the Mahalanobis distance on a subset of quantitative, ordinal and/or binary variables that can be conjointly considered as quasi-identifiers. Candidate nominal variables can be included after dichotomization. By using the Mahalanobis distance rather than other distance measures, we prevent highly dispersed and/or correlated groups of variables from exerting a preponderant effect on the distance, disregarding their significance in terms of identification disclosure risk. Moreover, the properties of the Mahalanobis distance are especially well understood with variables following normal distributions (75), which is generally the case with health data (76). However, atypical situations may arise where other distances would make unquestionably better choices. In such cases, we may use distance measures other than the Mahalanobis distance. This would for example be the case with data exclusively composed of binary variables, where we would tend to use the Jaccard index rather than the Mahalanobis distance. If the input data contain missing values and the distance measure selected does not support them, we advocate (and use) a conservative approach where:

- Distances between original units are computed after replacing missing values with values maximizing those distances (for example, if two units have a missing value for the same variable, we would assign the maximum value of the variable to the first unit, and the minimum value to the second).
- Distances between synthetic and original units are computed after replacing missing values with values minimizing those distances (using the same example as above, we would assign the exact same value of the variable to both units).

Following this approach can be seen as implementing a "worst-case scenario", because it causes overestimation of the distances between original units, and underestimation of the distances between original and synthetic units. When using this solution, it is thus hard for a synthetic unit to be retained, giving the guarantee that the resulting data preserve the confidentiality of original units at least as much as when no values are missing.

The second and third steps of the algorithm are the respective determination of the nearest original neighbor of each original and synthetic unit using the aforementioned distance. Finally, synthetic units which are too close to their nearest original neighbor are deleted, where "too close" means closer than the distance separating this original unit and its nearest original neighbor. New units are then synthetized using CART and undergo the same filtering process, until the desired number of retained units has been reached. Naive pseudocode for the whole process is given in Algorithm 1.

```
1  PROCEDURE filterSynthData:
2      origData, synthData
3  toFilter ← [ ]
4  FOR synthRow IN synthData :
5      minDistance ← −1
6      closestOrig ← nullRow
7      FOR origRow IN origData :
8          distance ← computeDist(synthRow, origRow)
9          IF minDistance = -1 OR distance < minDistance THEN
10             minDistance ← distance
11             closestOrig ← origRow
12     minDistOrig ← −1
13     FOR origRow IN origData :
14         distance ← computeDist(closestOrig, origRow)
15         IF (minDistOrig = -1 OR distance < minDistOrig) AND
            closestOrig ≠ origRow THEN
16             minDistOrig ← distance
17     IF minDistance < minDistOrig THEN
18         append(toFilter, synthRow)
19 removeRows(synthData, toFilter)
```

*Algorithm 1. Naive implementation of the distance-based filtering.*

This method has an intuitive appeal because it ensures that for every synthetic unit, the remark that it is close to an original one remains inconclusive: one can hardly expect synthetic units to be more different from the original units than the original units are from each other. If a participant to a study worries than a synthetic row resembles himself, he could always be reassured by hearing that at least one real person in the same dataset is even more resemblant to him. Said differently, he could have made the same remark if he had seen the original dataset with his data specifically removed, in which case this remark would be absurd. Therefore, if resemblance there is, it cannot be a legitimate cause for concern. On a more statistical point of view, this method also has the advantage of not preventing synthetic units from being close to one another. Neither does it prevent them from being close to original units if original units themselves are densely regrouped.

The practical gains of using this method in terms of privacy enhancement will be assessed throughout this work using methods developed below.

### 4. Noise addition

Fully synthetic data are often said to be immune from identity disclosure and are supposed to limit risks of attribute disclosure (71). In our framework, these risks are further mitigated by the manual SDC and filtering process described in previous sections. Still, synthetic data may be vulnerable to Membership Disclosure Attacks (MDA), where an attacker tries to determine whether an individual belongs to a dataset or not (37). This is especially the case with synthesis methods that sample synthetic values from original ones. For example, suppose that a row of a synthetic dataset generated using such a method contains the value "15" for the variable "number of children". Also suppose that all data originate from women (this may be known to any attacker, for example if the data was collected as part of a study on endometrial cancers). There are presumably very few women in any source population with this parity, generally no more than one. In such a scenario, the mere presence of the value "15" in the synthetic dataset indicates that this particular woman was part of the original dataset. This can have harmful consequences, for example if the data was collected as part of an oncology study and the attacker works for an insurance company.

Synthesis methods using CART are prone to such risks because they perform synthesis by sampling from actual variables values (cf. supra). Manual SDC described in the previous section is a first step to control membership disclosure risk, and by our standards must ensure enough granularity for qualitative variables. However, we regard it as insufficient for quantitative variables, especially because not all subjects exposed to this risk are outliers. Imagine, for example, a study on chronic hepatitis B (CHB) among farmers in French Guiana, where the prevalence of this condition was close to 5% in the 2010s (77). Considering a farmer population of around 1,500 individuals (78), this gives a prevalence of CHB among Guyanese farmers of around 75 individuals. Because of the imbrication of CHB outcomes with those of obesity (79), it is plausible that biometric variables such as weight and height will be collected in such a study. But due to the small number of individuals involved, it is also plausible that most heights in this population will be unique up to the nearest centimeter. If the data collected in the study contain height and is synthesized as is with CART, simply knowing the height of an individual (even a non-outlier) can therefore lead to membership disclosure. Of course, this would be an extreme case, as the source population for this study would be especially small. But the argument can also hold for larger populations, especially when considering less obvious variables or combinations of variables.

In the statistical community, the information that an individual belongs to a particular dataset has been acknowledged as especially sensitive as early as 1989 (80). Rather than a specific target for attacks, it was mainly treated as a preliminary information that any attacker must possess to perform fruitful identity or attribute disclosure attacks (81,82). This concern is still relevant today (69) and, together with their intrinsic sensitivity, makes membership information crucial to protect. Unfortunately, there seems to be little guidance in the literature on how to perform this protection (83). This can be due to the fact that microdata publication has historically been proposed for census data, in which membership disclosure risks are maximal by definition (84). Traditional syntactic privacy models, such as *k-anonymity* may contribute to protect synthetic data against membership disclosure (85), but are poorly suited to quantitative variables (86). One of the only specific metrics to assess membership disclosure risk, *δ-presence*, presents the same limitations (87). More recently have been proposed metrics derived from machine learning techniques, such as the F1 score (37,88), but their formal grounding seems fragile, and their application involves arbitrary configuration decisions (such as the proportion of records used to train the model).

Perturbative methods are an intuitive choice for protecting quantitative variables against membership disclosure. In his seminal paper on CART for synthetic data generation (42), Reiter suggested using Kernel Density Estimation (KDE) rather than direct sampling from leaves to prevent real values from being disclosed. The *synthpop* package implements this mechanism (to be activated via an optional parameter). However, the choice of the range for the KDE is left to the discretion of the user, without any formal hint on the resulting membership disclosure risk. Authors suggesting the use of additive noise for SDC are not much more informative, merely advising that the variance of the noise should be proportional to the variance of the original variable to preserve covariances (25,44,89).

For these reasons, we chose to develop our own formal model to ensure enough protection against MDA. In its current version, this model is based on the addition of uncorrelated noise to each quantitative variable of the synthetic dataset (synthesized without using KDE). Its mechanism and formal justification are presented in the following paragraphs.

Suppose that we want to protect a sample of size $n$ drawn from an underlying population of size $N$ against MDA, where only one quantitative variable is collected for each sampled unit. Let $(s_k)_{k \in [\![1;n]\!]}$ be these collected values, and $(x_k)_{k \in [\![1;N]\!]}$ be the values of this quantitative variable in the underlying population, sorted in ascending order. Finally, let $M$ be a probabilistic noise-addition mechanism independent on the values of input variables and preserving their means, and let define the sequence $(r_k)_{k \in [\![1;n]\!]}$ by: $r_k = M(s_k)$ for all $k \in [\![1;n]\!]$. Our goal is to shape $M$ so that the risk of membership disclosure is lower than some user-defined threshold when the user releases $(r_k)_{k \in [\![1;n]\!]}$.

For this, we model a typical scenario for a membership disclosure attack as follows. Suppose that an attacker knows the public version $(r_k)_{k \in [\![1;n]\!]}$ of our sample, a particular value $x_a$ of $(x_k)_{k \in [\![1;N]\!]}$, a plausible continuous probability distribution $D$ for $(x_k)_{k \in [\![1;N]\!]}$, the detailed behavior of the mechanism $M$ and the value of $N$ and $n$. With these information on hand, the attacker would presumably be interested in the probability that $x_a$ belongs to $(s_k)_{k \in [\![1;n]\!]}$ given $(r_k)_{k \in [\![1;n]\!]}$. To estimate this probability, the most natural choice would be to use some variation of the following formal model:

- let $(S_k)_{k \in [\![1;n]\!]}$ be independent random variables identically distributed according to a mixture distribution such that for every $k$ in $[\![1;n]\!]$, $S_k = x_a$ with probability $\frac{1}{N}$ and $S_k \sim D$ otherwise;
- let $(B_k)_{k \in [\![1;n]\!]}$ be independent and identically distributed random variables representing the noise added to $(s_k)_{k \in [\![1;n]\!]}$ by $M$;
- then the probability of interest is given by: $P(\cup_{k=1}^n S_k = x_a \mid \cap_{k=1}^n S_k + B_k = r_k)$.

However, this value alone would generally be of little use to a potential attacker. Let indeed $f : \begin{cases} \mathbb{R} \to [0;1] \\ x \mapsto P(\cup_{k=1}^n S_k = x \mid \cap_{k=1}^n S_k + B_k = r_k) \end{cases}$ be the application giving such a probability for any potential value of the variable considered. One shows easily that the elements of $\text{Im}((r_k)_{k \in [\![1;n]\!]})$ are the arguments of the local maxima of $f$. Consequently, unless $x_a$ is such a value, there will always be an infinite number of alternative real numbers associated with higher values of $f$ than $x_a$. Therefore, any claim of the form "$x_a$ is certainly in the dataset because $f(x_a)$ is high" could be countered by remarking that numerous other values would make better candidates, much more than there are in $\text{Im}((s_k)_{k \in [\![1;n]\!]})$.

To remain convincing, the attacker must at least show that the value of $f$ at $x_a$ is higher than at its neighboring values (that is $x_{a-1}$ and $x_{a+1}$, which by hypothesis the attacker has no way to prove that they do not exist). This implies that there exists a number $k \in [\![1;n]\!]$ such that $x_{a-1} < r_k < x_{a+1}$, which for $N$ large enough is true if $\frac{x_a + x_{a-1}}{2} < r_k < \frac{x_a + x_{a+1}}{2}$ in virtue of Taylor's theorem. To show that, the attacker would presumably use his knowledge of $x_a$, $N$ and $D$ to get a plausible value for $x_{a-1}$ and $x_{a+1}$. But even if he succeeds, the question remains whether the high value of $f(x_a)$ (that is, the proximity of $x_a$ to a value of $(r_k)_{k \in [\![1;n]\!]}$) is significant compared to the potentially high values of $f(x_{a-1})$, $f(x_{a+1})$, $f(x_{a-2})$, etc. A necessary condition to show this is arguably to prove that the presence of a value of $(r_k)_{k \in [\![1;n]\!]}$ between $\frac{x_a + x_{a-1}}{2}$ and $\frac{x_a + x_{a+1}}{2}$ is most probably due to $x_a$ rather than other values of $(x_k)_{k \in [\![1;N]\!]}$. Said differently, the point is to prove that the probability that $x_a$ has been sampled is high if there exists such a value of $(r_k)_{k \in [\![1;n]\!]}$. We call this probability the Elemental Correct Attribution Probability (ECAP) for $x_a$. If the attacker determines a high value for the ECAP, he can confidently affirm that $x_a$ has probably been sampled because other plausible values of $(x_k)_{k \in [\![1;N]\!]}$ cannot explain this particular value of $(r_k)_{k \in [\![1;n]\!]}$ alone.

On the contrary, if he fails to do so, he exposes himself to the arguments mentioned above and the attack cannot be considered successful.

Let $D'$ be a probability distribution resulting from the restriction of $D$ to $\mathbb{R}\backslash[x_{a-1}; x_{a+1}]$. Let $(S'_k)_{k\in[\![1;n]\!]}$ be independent random variables identically distributed according to a mixture distribution such that for every $k$ in $[\![1;n]\!]$, $S'_k = x_a$ with probability $\frac{1}{N}$ and $S'_k \sim D'$ otherwise. Once the attacker has estimated a value for $x_{a-1}$ and $x_{a+1}$, $(S'_k)_{k\in[\![1;n]\!]}$ is a more legitimate choice than $(S_k)_{k\in[\![1;n]\!]}$ to model the situation. Finally, let $I_1 = \left[\frac{x_a+x_{a-1}}{2}; \frac{x_a+x_{a+1}}{2}\right]$ and $I_2 = \left[\frac{x_{a-1}-x_a}{2}; \frac{x_{a+1}-x_a}{2}\right]$. The ECAP for $x_a$ is then given by:

$$\text{ECAP}(x_a) = P\left(\bigcup_{k=1}^n S'_k = x_a \mid \bigcup_{k=1}^n S'_k + B_k \in I_1\right)$$

Using elementary probability calculus operations, it is easy to show that:

$$\text{ECAP}(x_a) = 1 - \frac{\left(\frac{N-1}{N}\right)^n - \left(P(S'_1 + B_1 \notin I_1) - \frac{P(B_1 \notin I_2)}{N}\right)^n}{1 - P(S'_1 + B_1 \notin I_1)^n}$$

*(Equation 1)*

This formulation makes it apparent that the ECAP for a given value only depends on the distribution of $(B_k)_{k\in[\![1;n]\!]}$ once the problem has been modeled by a correct choice of $N$, $D$, $x_a$, $x_{a-1}$ and $x_{a+1}$. This property of the ECAP is what makes it useful to choose such a distribution, hence the determination of plausible values for $x_{a-1}$ and $x_{a+1}$ is especially critical. The choice of $N$ and $D$ is no less important, but relates more to the current state of scientific knowledge than to statistical considerations.

To determine plausible values for $x_{a-1}$ and $x_{a+1}$, one intuitive possibility would be to estimate the Mean Successive Differences (MSD) in the source population of the sample. Keeping previous notations, the MSD for variable x can be defined as:

$$\text{MSD} = \frac{1}{N-1} \sum_{k=1}^{N-1} x_{k+1} - x_k$$

When $x_1$ and $x_N$ are known, this obviously simplifies to $\text{MSD} = \frac{x_N - x_1}{N-1}$. However, this approximation of $x_{a-1}$ and $x_{a+1}$ is far from being optimal, especially with tight distributions where it tends to underestimate $x_{a-1}$ and $x_{a+1}$ for values close to the tail and to overestimate them otherwise. A better alternative is to use the distribution of $(x_k)_{k\in[\![1;N]\!]}$, which is known by hypothesis (or at least its continuous approximation) to estimate $x_{a-1}$ and $x_{a+1}$. With the knowledge attributed by hypothesis to the attacker, the following estimates are arguably the best to hold:

$$x_{a-1} \approx \frac{1}{\sum_{k=2}^N P(S_{(k)} = x_a)} \sum_{k=2}^N P(S_{(k)} = x_a) E(S_{(k-1)} \mid S_{(k)} = x_a)$$

$$x_{a+1} \approx \frac{1}{\sum_{k=1}^{N-1} P(S_{(k)} = x_a)} \sum_{k=1}^{N-1} P(S_{(k)} = x_a) E(S_{(k+1)} \mid S_{(k)} = x_a)$$

Where $S_{(k)}$ represents the $k$th order statistic of a sample of $N$ values originating from the same distribution than $(S_k)_{k\in[\![1;n]\!]}$. This is the estimate we use to determine ECAP. Rather than computing conditional expectations, we approximate these values using a Monte-Carlo approach where we sample $N-1$ values multiple times from distribution $D$ and calculate the distances of $x_a$ with the values to which it is closest. We found this approach to be computationally inexpensive while generating precise estimates after a few iterations only.

Using these estimates, calculating the ECAP for a particular value of $(x_k)_{k \in [\![1;N]\!]}$ is straightforward. Depending on $D$ and on the desired type of distribution for $(B_k)_{k \in [\![1;n]\!]}$, this calculation may require classical Monte-Carlo techniques. Figure 5 shows the resulting ECAP for $x_a = 178$, $D = N(170, 12^2)$, $N = 1500$, $n = 25$ and $B_1 = N(0, \sigma^2)$ with various values of $\sigma^2$ and where $N(a, b^2)$ denotes the normal distribution of parameters $a$ and $b^2$. This would arguably correspond to the ECAP for the height of a 178 centimeters tall individual sampled from the population of Guyanese farmers.

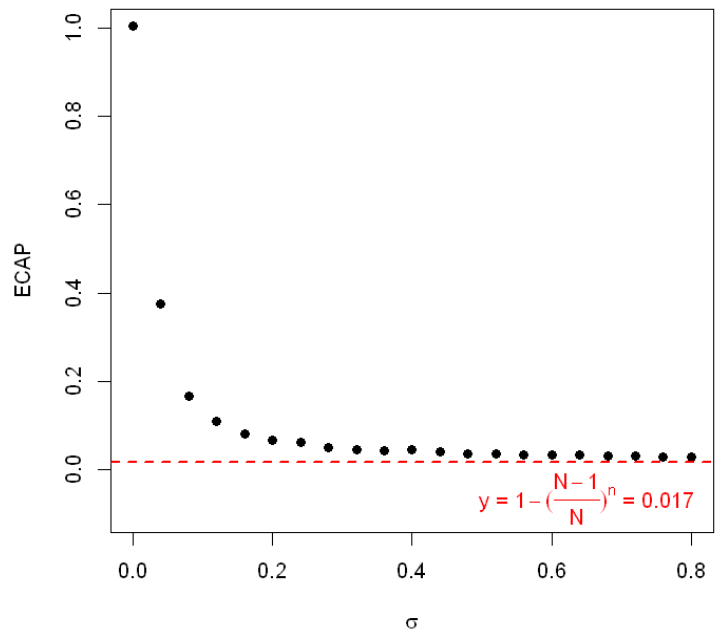

*Figure 5. ECAP values after the addition of normal noise of increasing variance.*

As expected, the calculated ECAP start at 1 (when $\sigma = 0$) and quickly drop close to $1 - \left(\frac{N-1}{N}\right)^n \approx 0.017$ with values larger than 0.2. The resulting curve has a steep initial slope, which is explained by the large value of $N$. Scatter plots of the ECAP for alternative values of $N$ are displayed in Figure 6.

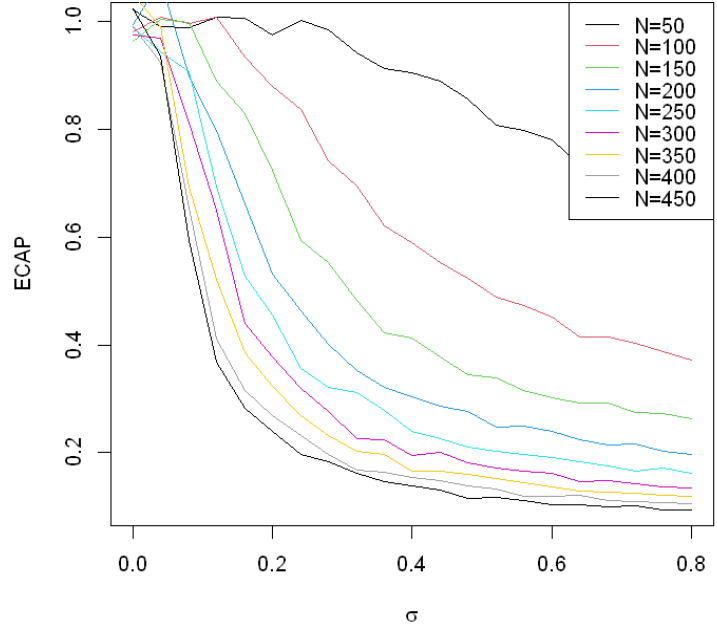

*Figure 6. ECAP values after the addition of normal noise of increasing variance for various population sizes.*

For populations large enough, Figure 6 shows that calculated ECAP with the same values of $x_a$, $n$ and $\sigma^2$ are quite insensitive to the exact value of $N$. As illustrated by Figure 7, the same holds with the variance of $D$ in the case of normal distributions. For populations large enough, the exact choice of $N$ and $D$ should therefore not be a major cause for concern.

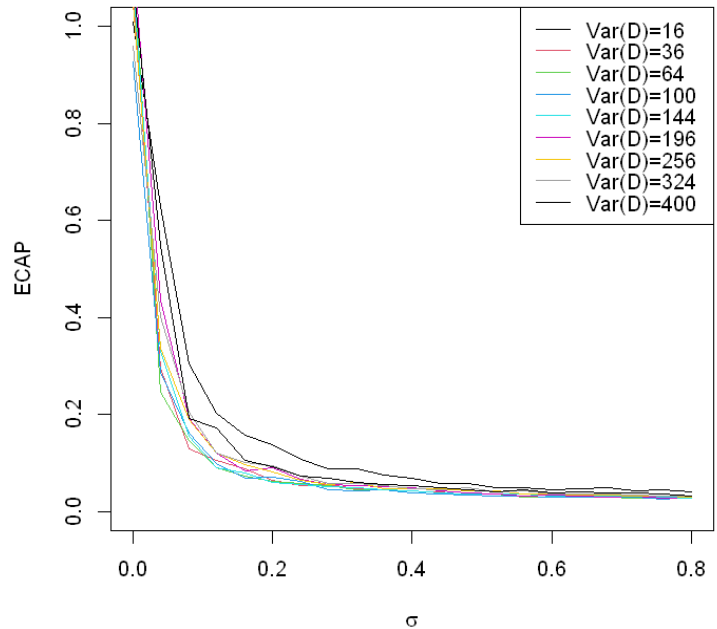

*Figure 7. ECAP values after the addition of normal noise of increasing variance for N=1500 and various population variances.*

Because of its high face validity, its robustness to changes in the underlying parameters, and its acceptable computational complexity, we think that the ECAP is a relevant metric for SDC against MDA. After correctly modelling the situation (which in most cases should not be a major challenge), data holders should be able to calculate ECAP for each observation of quantitative variables in their dataset. They can then choose a noise distribution making these ECAP low enough so that an attacker would not be able to demonstrate membership, even with an extensive knowledge of the variables involved and the type of noise used. In this process, a decision has to be made on the maximum ECAP to be tolerated, which is always arbitrary to some extent. Intuitively, ECAP values below 0.2 are suggestive of high protection against MDA, and determining the minimal noise needed to attain such values is straightforward using standard optimization algorithms. However, we suggest having a more flexible approach by looking at ECAP plots like the one in Figure 5 and selecting values based on the successive difference quotients. This may help minimizing the risk of membership disclosure without significantly impairing the utility of resulting data. For example, in Figure 5, an ECAP of value 0.2 would approximately be attained with $\sigma = 0.075$, but taking $\sigma = 0.1$ instead would ensure an ECAP of around 0.1. Said differently, a difference of only 0.025 in the noise variance would lead to a 50% reduction in the ECAP. In this case, we would recommend retaining the value 0.1, which is what we typically do in our synthesis framework. Finally, the question remains as to which value to choose for $x_a$. To ensure a sufficient protection of all original units, we suggest computing the ECAP for all available values and choosing a noise distribution minimizing the highest ECAP. The corresponding original value will generally be close to the tails of the distribution, as these values typically exhibit the highest distances with neighboring values in the population.

Depending on the parameters used, this framework can allow for effective protection against MDA without significantly compromising data utility. For example, in a random sample of 15 heights drawn from the population of Guyanese farmers (see above for the model specifications), obtaining ECAP below 0.1 for all subjects only requires adding a normal noise with a variance of 0.09. Figures 8 and 9 illustrate the resulting perturbation, which is arguably minimal. Consequently, we perform these operations for every quantitative variable of our synthetic datasets that need to be protected against MDA. Because needed variances are often small, they rarely result in aberrant values (such as negative heights), even though we use probability distributions defined on the whole real line. Of course, only synthesized values are protected this way, as values that have not been sampled by CART present by definition no risk of disclosure. Finally, we recommend publishing the type and amount of noise added to these variables so that end-users of the data can take this information into account for their fine-grained analyses. This presents no risk of disclosure, as this information is by hypothesis part of the knowledge of a potential attacker (cf. supra). However, we strongly discourage against publishing obtained ECAP values, as this could lead to privacy breaches if an attacker uses a reversed version of Equation 1.

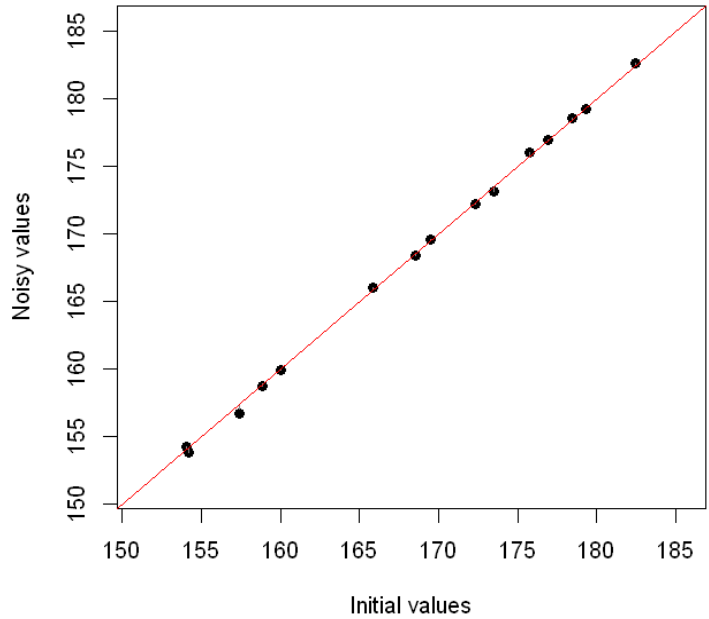

*Figure 8. Deviation from initial position after the addition of a normal noise of 0.09 variance to a random sample of heights.*

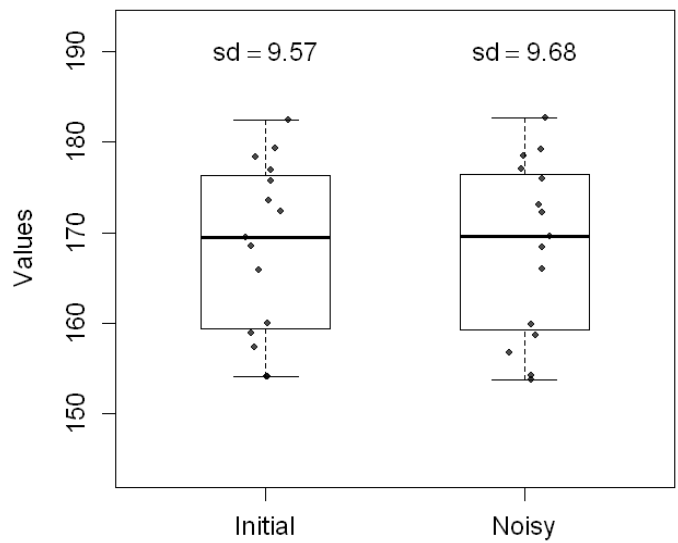

*Figure 9. Distribution of the sample displayed in Figure 8 before (left) and after (right) noise addition.*

### Assessment framework

To assess the performance of our SDGF in terms of privacy and utility of the data generated, we computed various metrics on the original and synthetic version of a rich dataset. We also performed similar evaluations on datasets generated using truncated versions of our framework to validate the individual steps involved. In the following sections, we successively describe the original dataset used in this process, the synthetic versions we generated from it and the privacy and utility assessment framework we used for their evaluation.

## *1. Original dataset*

The dataset we used to assess the performance of our SDGF originates from the "Mental Health In Prison" study conducted by Falissard *et al.* between 2003 and 2004 to determine the prevalence and risk factors of mental disorders in French prisons (90). This epidemiological study used a two-stage stratified random sampling strategy. At first, 20 prisons were selected at random from the list of all French metropolitan prisons for men with stratification on the type of prison. Prisoners were then chosen at random in each of these 20 prisons until 799 prisoners were enrolled. Each prisoner was interviewed for approximately 2 hours by a group of 2 clinicians, one of whom was called a "senior" clinician and the other a "junior". These clinicians independently established their positive and severity diagnoses which they then pooled together. This resulted in a dataset of 799 observations of several hundreds of variables, of which we selected 26 based on their theoretical and empirical importance. Table 2 summarizes the nature of these variables after classical data management operations and manual SDC. In the following, we will refer to this original dataset as $D_I$.

| Variable name | Meaning | Type and values |
|---|---|---|
| AGE | Age in years | Quantitative (positive integers) |
| JOB | Job of the prisoner | Nominal:<br> -0: farmer<br> -1: craftsman<br> -2: manager<br> -3: intermediate<br> -4: employee<br> -5: worker<br> -6: other<br> -7: no job |
| DURATION | Length of incarceration | Ordinal:<br> -1: less than a month<br> -2: 1 to 6 months<br> -3: 6 to 12 months<br> -4: 1 to 5 years<br> -5: more than 5 years |
| DISCIPLINARY | History of disciplinary action in prison during the current incarceration | Binary:<br> -0: no<br> -1: yes |
| N.CHILDREN | Number of children | Quantitative (positive integers) |
| N.SIBLINGS | Number of siblings | Quantitative (positive integers) |
| EDUCATION | Level of education | Ordinal:<br> -1: no diploma<br> -2: middle school diploma<br> -3: vocational certificate<br> -4: high school diploma<br> -5: university diploma |
| SEPARATION | History of separation from at least one of the parents during at least 6 months | Binary:<br> -0: no<br> -1: yes |
| CHILDREN.JUDGE | History of supervision by a juvenile court judge before the age of 18 | Binary:<br> -0: no<br> -1: yes |
| PLACEMENT | History of placement measures (group | Binary: |
| | home, foster family, etc.) | -0: no<br>-1: yes |
| ABUSE | History of childhood abuse (of physical, psychological, or sexual nature) | Binary:<br>-0: no<br>-1: yes |
| SEVERITY | Overall severity of the prisoner's disorders as assessed by both clinicians | Ordinal:<br>-1: normal<br>-2: borderline<br>-3: mild<br>-4: moderate<br>-5: marked<br>-6: severe<br>-7: among the most ill patients |
| DEPRESSION | Existence of a depressive disorder (based on the consensus of the two clinicians) | Binary:<br>-0: no<br>-1: yes |
| AGORAPHOBIA | Existence of an agoraphobic disorder (based on the consensus of the two clinicians) | Binary:<br>-0: no<br>-1: yes |
| PTSD | Existence of a post-traumatic stress disease (based on the consensus of the two clinicians) | Binary:<br>-0: no<br>-1: yes |
| ALCOHOL | Existence of an alcohol abuse (based on the consensus of the two clinicians) | Binary:<br>-0: no<br>-1: yes |
| SUBSTANCE | Existence of a substance abuse (based on the consensus of the two clinicians) | Binary:<br>-0: no<br>-1: yes |
| SCHIZOPHRENIA | Existence of a schizophrenia (based on the consensus of the two clinicians) | Binary:<br>-0: no<br>-1: yes |
| PERSONALITY | Intensity of a potential personality disorder, related to the Temperament and Character Inventory (TCI) by R. Cloninger (91) | Ordinal:<br>-1: absent<br>-2: mild<br>-3: moderate<br>-4: severe |
| NS | Novelty seeking as defined in the TCI by R. Cloninger (91) | Ordinal:<br>-1: low<br>-2: moderate<br>-3: high |
| HA | Harm Avoidance as defined in the TCI by R. Cloninger (91) | Ordinal:<br>-1: low<br>-2: moderate<br>-3: high |
| RD | Reward Dependence as defined in the TCI by R. Cloninger (91) | Ordinal:<br>-1: low<br>-2: moderate<br>-3: high |
| SUICIDE.SCORE | Aggregated suicide risk score | Ordinal:<br>-1: low<br>-2 to 5: increasing severity<br>-6: severe |
| SUICIDE.HR | Existence of a high risk of suicide | Binary: |

| | | -0: no |
| | | -1: yes |
| SUICIDE.PAST | History of at least one past suicide attempt | Binary: <br> -0: no <br> -1: yes |
| DUR.INTERV | Duration of the interview in minutes | Quantitative (positive integers) |

*Table 2. Characteristics of variables in the original dataset.*

## 2. Generated datasets

Because some components of our privacy and utility assessment frameworks require a control dataset that has not been used in the synthesis processed, we did not use all rows of $D_I$ as input to our SDGF. Instead, we randomly selected 199 rows of $D_I$, which we stored in a separated dataset called $C$ intended to be used as a control dataset when needed. The remaining 600 rows of $D_I$ (which we shall call $D_O$ in what follows) were used to generate 3 synthetic datasets of same length, which we respectively refer to as $D_1$, $D_2$ and $D_3$. Each of them corresponds to a further step of our SDGF. As such, $D_1$ was generated using CART implemented in the R 4.3.0 *synthpop* package and running with *rpart*. The CART were configured so that each leaf contained at least 33 observations, and that splits were only attempted in parent nodes containing at least 100 observations. We anticipated that these parameters would provide a satisfactory level of disclosure control given the nature of the original data.

$D_2$ was generated from $D_1$ by running our distance-filtering algorithm described above. Because most of the variables involved were quantitative, binary or ordinal, we chose the Mahalanobis distance as the measure for filtering rows. We intended to use all variables as identifiers, except *job* which was the only nominal variable. Missing data were handled using our custom conservative method based on worst-case scenarios (see above).

Finally, we generated $D_3$ by adding noise to $D_2$'s quantitative variables that needed formal protection against MDA. Based on official reports, we estimated that the total number of males held in French prisons was about 60,000 in 2003 and 2004, at the time the data were collected (92–94). These reports and some other publications allowed us to gain direct or indirect information on the distribution of these quantitative variables in the source population of the study (90,95–100). More specifically, we considered that the following distributions hold in this source population: $\text{AGE} \sim N(34, 8^2)$, $\text{N.CHILDREN} \sim N(2, 2.5^2)$ and $\text{N.SIBLINGS} \sim N(4, 2.5^2)$. The only other quantitative variable in the dataset, DUR.INTERV, did not undergone noise addition as per definition it cannot be collected secondarily and thus cannot be subject to MDA. In the end, $D_3$ is the synthetic dataset we would typically want to release as part of the Open-CESP initiative.

## 3. Privacy-related assessment

The ability of synthetic data generation methods to ensure the privacy of individuals from which their input data originate is considered to be one of their most essential qualities. It is typically assessed using privacy metrics measuring different aspects of the protection they provide against various attacks (34). Many metrics have been proposed, mainly targeting partially synthetic data which can be exposed to the same attacks than non-synthetic data (101). On the other hand, fully synthetic data generation methods such as ours are virtually unexposed to most of these attacks, including identity disclosure attacks (made ineffective by the synthesis process and the subsequent filtering) and MDA (formally prevented by the noise addition). They can, however, be subject to attribute disclosure attacks, where an attacker tries to infer the value of target variables for an individual based on his knowledge of other "key" variables (71).

Quantifying the information that an attacker can gain conditionally on his knowledge of key variables is a challenging task for which no formal model currently exists. In 2014, Mark Elliot made a valuable contribution to this field by introducing a new metric initially named the "Empirical differential privacy" (102) and developed in subsequent works under the name of Differential Correct Attribution Probability (DCAP) (103). Given a set $K$ of key values and a set $T$ of target values, the idea is to compare the probability $P(T \mid K)$ in the original and synthetic datasets. This is typically implemented by counting the number of occurrences of these values in each dataset. A synthetic dataset giving a conditional probability close to the one calculated in the original data would be considered as particularly disclosive. On the contrary, if the probability obtained from the synthetic data is close to the univariate distribution of the target variables, then the risk of attribute disclosure can be considered as minimal.

The DCAP has several qualities, including its high face validity upon initial approach and the possibility to compute it using specifically chosen key and target variables, thus making it easier to interpret. However, it faced a serious criticism in 2019, when Chen *et al.* showed that in its original state, the DCAP was no more than a utility metric (104). Indeed, let us imagine a key variable consisting of individuals' smoking status and a target variable defined as whether or not they have lung cancer. If a synthetic dataset allowed a precise prediction of the oncological status of individuals conditionally on their smoking habits, would this count as disclosure or rather as a sign of quality of the synthetic data? Arguably, preventing synthetic datasets from drawing such inferences would defeat their whole purpose, even with other choices of key and target variables. The correction suggested by Chen *et al.* was already outlined in the seminal works on the DCAP, and consisted in restricting the calculation of conditional probabilities on statistical uniques. The resulting metric has been called the Targeted Correct Attribution Probability (TCAP). Because it intuitively resolves the main flaws of the DCAP, this new metric has enjoyed some popularity since then, and has been used in several other works (105–107). Still, it has the disadvantage of being only defined for categorical variables, and thus unsuitable to rich datasets like those processed by the Open-CESP team. In 2019, Hittmeir *et al.* suggested a way of extending the DCAP to numerical variables, which could theoretically be adapted to the TCAP (61). However, their approach suffers from several flaws that we consider to be prohibitive, namely their choice of an infinite radius to compare the value of numerical variables, which raises interpretation issues. They also seem to recode categorical variables into numerical, which in our view would make their approach as unsuitable to the Open-CESP as the original TCAP.

Even though the TCAP and its variants are not suited to our privacy assessment framework in their current state, we think that their general approach was the most promising to quantify attribute disclosure risks. Consequently, we chose to design a new metric following the same core idea, but suited both to numerical and categorical variables, which we call the Generalized Targeted Correct Attribution Probability (GTCAP). Our approach is based on the specification of radiuses for all quantitative variables involved. These radiuses define the maximum distance between two values of these variables so that they can be considered equal or approximately equal. More specifically, they allow for the computation of proximities between each value of these variables, linearized such that they equal 0 for an absolute difference greater than the radius, and 1 for a null difference. These proximities are used both to define statistical uniques and to weight the conditional probabilities computed from them.

When no numerical variables are involved, or when every computed proximities are either 0 or 1, these conditional probabilities are the same that would be obtained by computing a TCAP on the same dataset with categorized variables. When the calculation of a conditional probability involves a null denominator, we follow one of the option proposed by Taub *et al.* and count the resulting correct attribution probability as 0 (103). Finally, we normalize resulting GTCAP between the univariate probability and the GTCAP obtained with the original data. Normalized GTCAP obtained from all statistical uniques are then averaged, resulting in what we call the mean GTCAP for a synthetic dataset. Algorithm 2 contains the key elements of our implementation.

```
1  FUNCTION getTargetsProxCoef:
2  |  row1, row2, targets, targetsR
3  targets1Val ← filter(row1, targets)
4  targets2Val ← filter(row2, targets)
5  catTargets1Val ← filterCat(targets1Val)
6  catTargets2Val ← filterCat(targets2Val)
7  numTargets1Val ← filterNum(targets1Val)
8  numTargets2Val ← filterNum(targets2Val)
9  IF catTargets1Val ≠ catTargets2Val THEN
10 |  RETURN 0
11 ELSE IF length(numTargets1Val) = 0 THEN
12 |  RETURN 1
13 ELSE
14 |  diff ← abs(numTargets1Val − numTargets2Val)
15 |  RETURN sum(pmax(0, 1 − diff/targetsR)) /
       length(numTargets1Val)

16 FUNCTION getKeysProxCoef:
17 |  row1, row2, keys, keysR
18 keys1Val ← filter(row1, keys)
19 keys2Val ← filter(row2, keys)
20 catKeys1Val ← filterCat(keys1Val)
21 catKeys2Val ← filterCat(keys2Val)
22 numKeys1Val ← filterNum(keys1Val)
23 numKeys2Val ← filterNum(keys2Val)
24 IF catKeys1Val ≠ catKeys2Val THEN
25 |  RETURN 0
26 ELSE IF length(numKeys1Val) = 0 THEN
27 |  RETURN 1
28 ELSE
29 |  diff ← abs(numKeys1Val − numKeys2Val)
30 |  RETURN sum(pmax(0, 1 − diff/keysR)) /
       length(numKeys1Val)

31 FUNCTION computeGTCAP:
32 |  data, row, keys, keysR, targets, targetsR
33 keysCoefs ← [ ]
34 targetsCoefs ← [ ]
35 FOR dataRow IN data :
36 |  append(keysCoefs, getKeysProxCoef(row, dataRow,
       keys, keysR))
37 |  append(targetsCoefs, getTargetsProxCoef(row,
       dataRow, targets, targetsR))

38 IF sum(keysCoefs) = 0 THEN
39 |  RETURN 0
40 ELSE
41 |  RETURN sum(keysCoefs ∗ targetsCoefs) /
       sum(keysCoefs)

42 FUNCTION univariatePrediction:
43 |  data, row, targets, targetsR
44 targetsRowVal ← filter(row, targets)
45 catTargetsRowVal ← filterCat(targetsRowVal)
46 numTargetsRowVal ← filterNum(targetsRowVal)
47 correctness ← [ ]
48 FOR dataRow IN data :
49 |  targetsDataRowVal ← filter(dataRow, targets)
50 |  catTargetsDataRowVal ←
       filterCat(targetsDataRowVal)
51 |  numTargetsDataRowVal ←
       filterNum(targetsDataRowVal)
52 |  IF catTargetsRowVal ≠ catTargetsDataRowVal
       THEN
53 |  |  append(correctness, 0)
54 |  ELSE IF length(numTargetsRowVal) = 0 THEN
55 |  |  append(correctness, 1)
56 |  ELSE
57 |  |  diff ← abs(numTargetsRowVal −
          numTargetsDataRowVal)
58 |  |  append(correctness, sum(pmax(0, 1 −
          diff/targetsR)) / length(numTargetsRowVal))
59 RETURN mean(correctness)

60 FUNCTION computeMeanGTCAP:
61 |  origData, synthData, keys, keysR, targets, targetsR
62 uniqueRows ← getUniqueRows(origData, keys, keysR,
     targets, targetsR)
63 res ← [ ]
64 FOR row IN uniqueRows :
65 |  synthVal ← computeGTCAP(synthData, row, keys,
       keysR, targets, targetsR)
66 |  baseVal ← univariatePrediction(origData, row,
       targets, targetsR)
67 |  origVal ← computeGTCAP(origData, row, keys,
       keysR, targets, targetsR)
68 |  append(res, normalize(synthVal, baseVal, origVal))
69 RETURN mean(res)
```

*Algorithm 2. Main functions needed to compute the mean GTCAP of a synthetic dataset.*

To assess the performance of our SDGF in terms of protection against attribute disclosure attacks, we computed the mean GTCAP for $S_1$, $S_2$ and $S_3$ using AGE, JOB, N.CHILDREN, N.SIBLINGS and EDUCATION as key variables, and SEVERITY as target variable. The decision to include only one target variable was made to prevent computed probabilities from being zero too frequently. The number of key variables was also limited so that only an acceptable proportion of units would become statistical uniques. The following radiuses were used: 5 for AGE, 1 for N.CHILDREN and 2 for N.SIBLINGS. Additionally to the mean GTCAP, we also used histograms to display the normalized GTCAP of all statistical uniques and assess their dispersion for each synthetic dataset.

Although their historical usage was mainly related to the assessment of the privacy of synthetic datasets used by computer scientists to train artificial intelligence models (32), metrics derived from machine learning techniques are expanding in the field of synthetic health data generation (108). Among them, those provided by the Anonymeter framework have a particular appeal because of their apparent regulatory validity. Running on Python 3.8 to 3.10, Anonymeter provides three main classes, each assessing a different type of disclosure risks, respectively singling out risks, linkability risks and inference risks (109). This typology of risks is based on the semantics used by the former European Union Data Protection Working Party (110). The assessment framework itself is based on simulated attacks exploiting each of these risks. According to its authors, this framework has been positively reviewed by the French Commission Nationale de l'Informatique et des Libertés (CNIL), which has "not identified any reason suggesting that the proposed set of methods could not allow to effectively evaluate the extent to which the aforementioned three criteria are fulfilled or not in the context of production and use of synthetic datasets" (111). Because of this apparent legal compliance and its complementarity with more standard metrics, we also included Anonymeter in our privacy assessment framework. We ran it on Python 3.10 to simulate 150 univariate and multivariate singling-out attacks as well as 150 linkage attacks against the following groups of variables: AGE, JOB, EDUCATION, N.CHILDREN and N.SIBLINGS versus DURATION, DISCIPLINARY, CHILDREN.JUDGE and PLACEMENT. Finally, we used it to simulate 26×150 inference attacks successively targeting each variable of our synthetic datasets, with all other variables being used as keys. In all these simulations, $C$ was used as the learning dataset and $S_1$, $S_2$ and $S_3$ were the successive targets.

### 4. Utility-related assessment

Along with privacy, utility of synthetic data is one of their main sought qualities. Although some authors may use this term with a more specific meaning, it can be defined as the ability of the synthetic data to replace the original ones in their intended use (34). In the context of the Open-CESP initiative, one of the main component of the utility of synthetic data should therefore be their ability to replicate the statistical content of the original data. This arguably implies a high distributional similarity between the synthetic and original data. The first step of our utility assessment framework was therefore to compare the univariate distribution of all variables in $D_O$ and (respectively) $S_1$, $S_2$ and $S_3$. To compare bivariate distributions between each of these datasets, we also plotted scatter plots of every pair of their quantitative variables.

The intended use of research data can be considered from two perspectives. On the one hand, the specific context of the research can suggest specific processing and analyses of the data. On the other hand, the research dataset may virtually be processed in a infinite number of manners, each of them having an equal a priori validity. In the synthetic data literature, both these perspectives have been taken to justify two different types of utility metrics: narrow metrics and broad ones (48). Narrow metrics are by definition specific to a particular analysis to be performed on the data; often, they simply consist in the replication of such an analysis.

In contrast, utility metrics are said to be broad when they aim at quantifying the overall utility of a synthetic dataset, irrespective of the analyses to be performed on it.

In the absence of formal ways to prove the optimality of a SDGF in terms of utility of the generated data, the use of metrics of one type or the other is considered to be a critical part of any utility assessment framework. In this context, narrow and broad metrics are not incompatible and some authors even suggest that they should be used together when possible (49). The large number of reported metrics in the literature could make the choice of a specific metric difficult. However, when validating a novel SDGF, we believe that already proven metrics are the most suitable, as it would be otherwise difficult to validate separately the SDGF and the metric. For this reason, we chose to use the so-called Purdam and Elliot methodology (112) for the narrow utility assessment of our synthetic data, and the propensity score Mean Squared Error (pMSE) as a broad utility metric.

The Purdam and Elliot methodology has been first described by Taub *et al.* in 2017 (113), referring to a previous work by Purdam *et al.* (112). Given an original dataset collected as part of a study, the core of this method is to identify the essential findings of the original study and to try to replicate them with the synthetic data. Depending on the statistical nature of these findings, specific metrics can be calculated such as interval overlaps or ratios of counts (113). In our case, the main findings associated with the original data have been published by Falissard *et al.* in 2006 (90). In this article, the most crucial data are arguably shown in the first table, which contains the prevalence estimates of several DSM-IV diagnoses. Using $D_O$, $S_1$, $S_2$ and $S_3$, we would be able to calculate similar estimates for major depressive disorders, agoraphobia, substance-related disorders and schizophrenia, which is the way we chose to implement the Purdam and Elliot methodology in our assessment framework.

As for the pMSE, it has been extensively studied in recent works (114) and has even been advocated as the best and most popular population-level utility metric (115). As its name suggests, it is based on the so-called propensity scores used in causal inference studies. The idea is to use all available variables to build a model predicting the synthetic nature of a given row (original or synthetic). For each row, this predicted probability is subtracted from the real proportion of synthetic rows in the whole dataset combining original and synthetic units. Each differences are squared and their mean gives the pMSE. Intuitively, a low pMSE indicates that the model was not able to distinguish correctly between original and synthetic units, thus suggesting a high utility of synthetic data (116). In our utility assessment framework, we used the functions provided by the *synthpop* package to calculate the pMSE for $S_1$, $S_2$ and $S_3$ based on a propensity model fitted using CART (117). For each synthetic dataset, we also plotted the standardized pMSE ratios obtained from each pair of variables. Standardized pMSE ratios can be interpreted as classical pMSE, but allow meaningful comparisons between several models. Their calculation method is detailed in (116).

# Results

All of the following results were obtained in a single pass, using a random seed unknown to the authors.

## Synthesis

The synthesis of $S_2$ from $S_1$ was first attempted using all non-ordinal variables to calculate distances, as described in the "Methods" section. By doing so, however, the resulting covariance matrix was found to be near-singular, having a determinant of about $1.8 \times 10^{-7}$ which resulted in numerical stability issues when computing Mahalanobis distances. We therefore decided to remove the variables most likely involved in this near-singularity from the distance calculations. Figure 10 shows the initial correlation matrix between all non-ordinal variables of $D_O$. From the inspection of this matrix, it appeared that SUICIDE.HR, SUICIDE.SCORE, CHILDREN.JUDGE and PERSONALITY were responsible for diffuse correlation patterns. Accordingly, we exclude them from the calculation of Mahalanobis distances. The determinant of the covariance matrix without these variables had a satisfactory value of around $0.0002$, allowing the synthesis of $S_2$ from $S_1$ without further issues. Figure 11 shows the correlation matrix of $D_O$ without these variables. Figures 12 shows a projection performed using NIPALS (118) of the units synthesized and filtered after the first iteration. Figures 13 uses the same projection to display retained synthetic units.

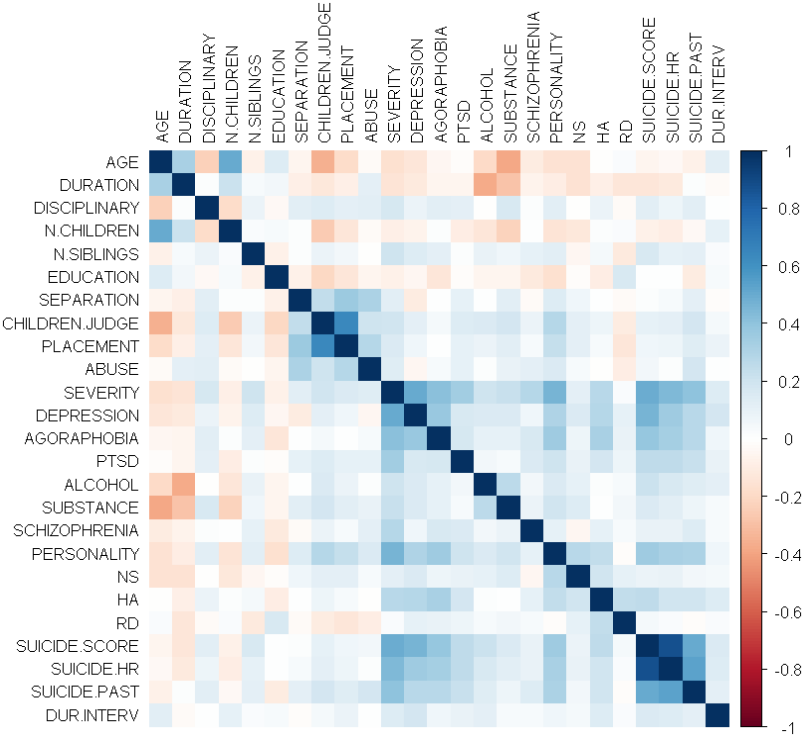

*Figure 10. Correlation matrix for all non-ordinal variables of $D_o$.*

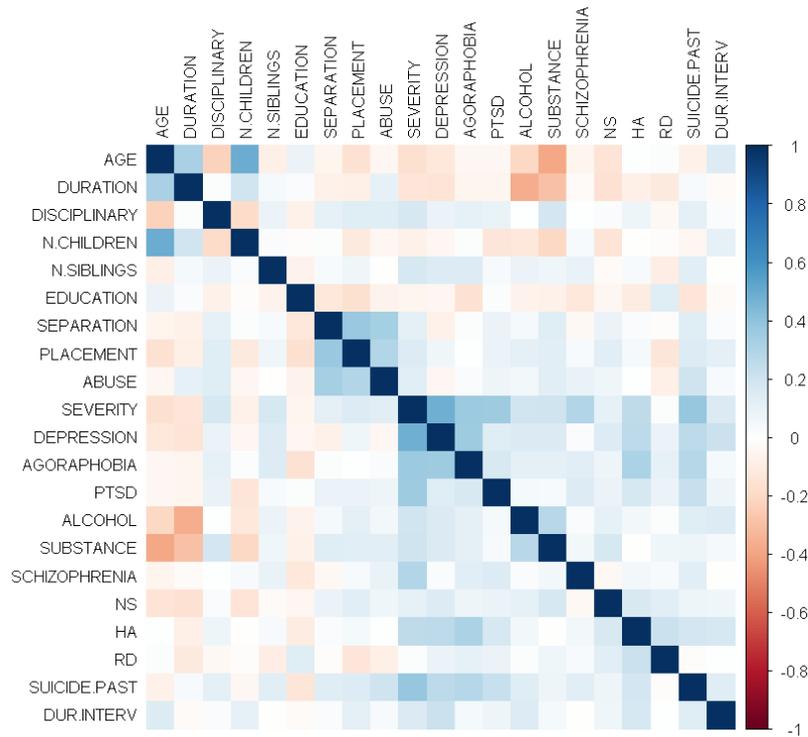

*Figure 11. Correlation matrix for all non-ordinal variables of $D_o$, excluding SUICIDE.HR, SUICIDE.SCORE, CHILDREN.JUDGE and PERSONALITY.*

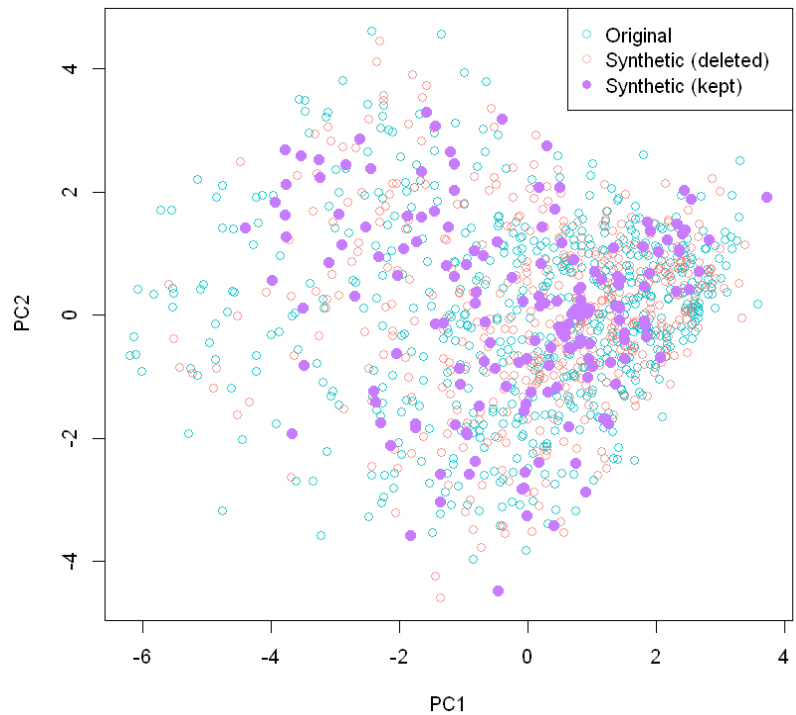

*Figure 12. Projection of $D_o$ and S1 on their first (PC1) and second (PC2) principal component. Points of $S_1$ are displayed red if they were filtered out, and purple otherwise.*

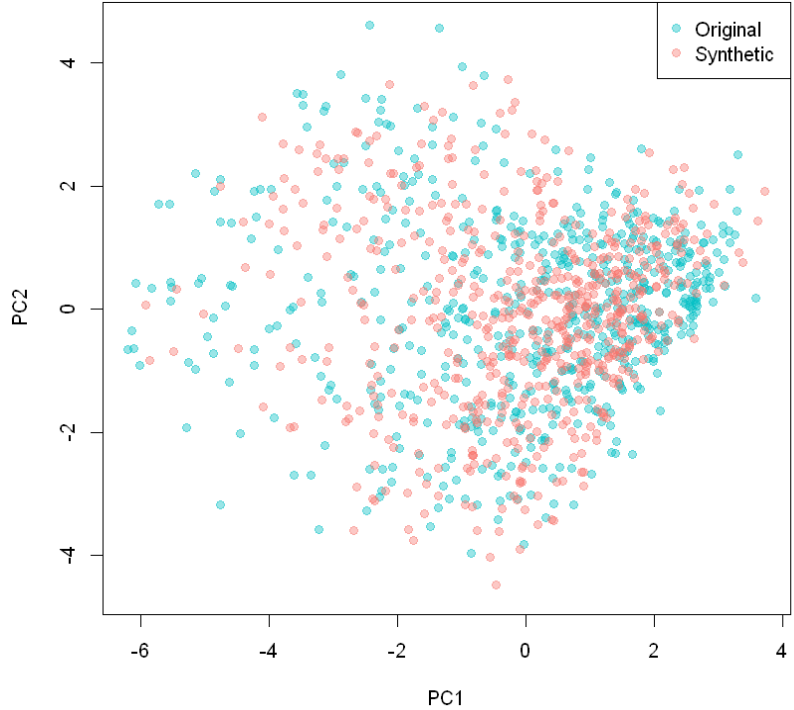

*Figure 13. Projection of $D_o$ and $S_2$ on the same components as in Figure 12. Points of $S_2$ do not appear to be further from $D_o$ than those of $S_1$.*

Based on the model specified above and the value of the synthesized units for numerical variables, we determined that a normal noise with a standard deviation of 0.1 was needed to protect both AGE and N.CHILDREN against MDA, while a standard deviation of 0.2 was needed to protect N.SIBLINGS. $S_3$ was obtained from $S_2$ by applying these additive noises.

**Privacy-related assessment**

The mean GTCAP for $S_1$, $S_2$ and $S_3$ were respectively 0.183, 0.164 and 0.166, suggesting an excellent level of protection against attribute disclosure attacks. Histograms of normalized GTCAP for statistical uniques can be seen in Figure 14.

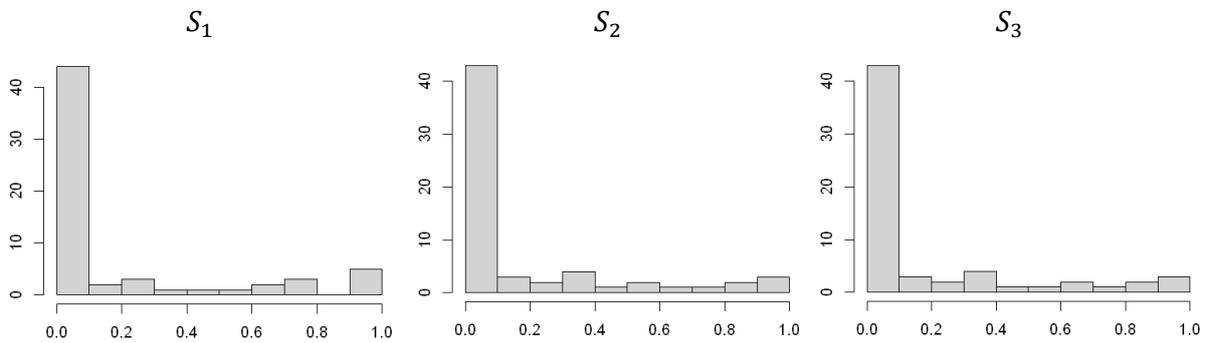

*Figure 14. Normalized GTCAP values for statistical uniques in $S_1$, $S_2$ and $S_3$.*

The results of attacks simulated using Anonymeter are displayed in Table 3 and Figures 15 to 17. $S_2$ and $S_3$ seemed to perform better than $S_1$ for all attack types, which was expected given the additional processing that these two datasets underwent. Quantifying the relative performance of the three datasets was, however, tedious because of the imprecision of the score estimates. This was especially apparent when comparing the scores obtained by $S_2$ and $S_3$, which were significantly different despite these datasets being constituted from almost the same rows. The precision could anyways not have been a lot better because of the necessarily limited size of the control dataset.

| Attack type | $S_1$ | $S_2$ | $S_3$ |
|---|---|---|---|
| Singling out / univariate | 0.200 | 0.226 | 0.206 |
| Singling out / multivariate | 0.242 | 0.210 | 0.084 |
| Linkability | 0.006 | 0.000 | 0.000 |

Table 3. Scores of each synthetic dataset after simulated attacks by Anonymeter. Lower scores are indicative of a better protection against these type of attacks.

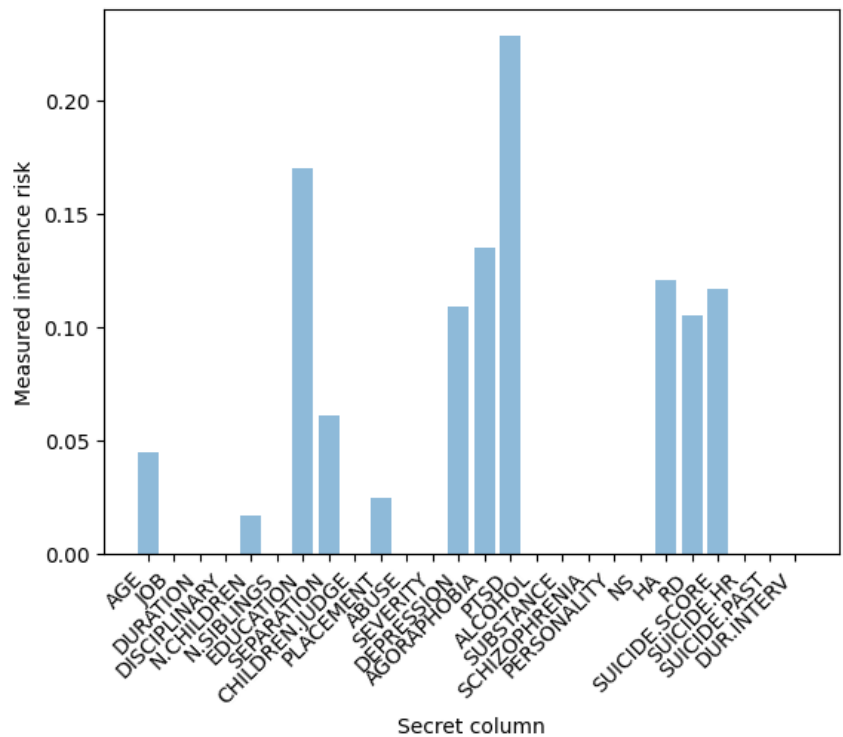

Figure 15. Scores obtained by $S_1$ after simulated inference attacks with each variable being successively selected as a target.

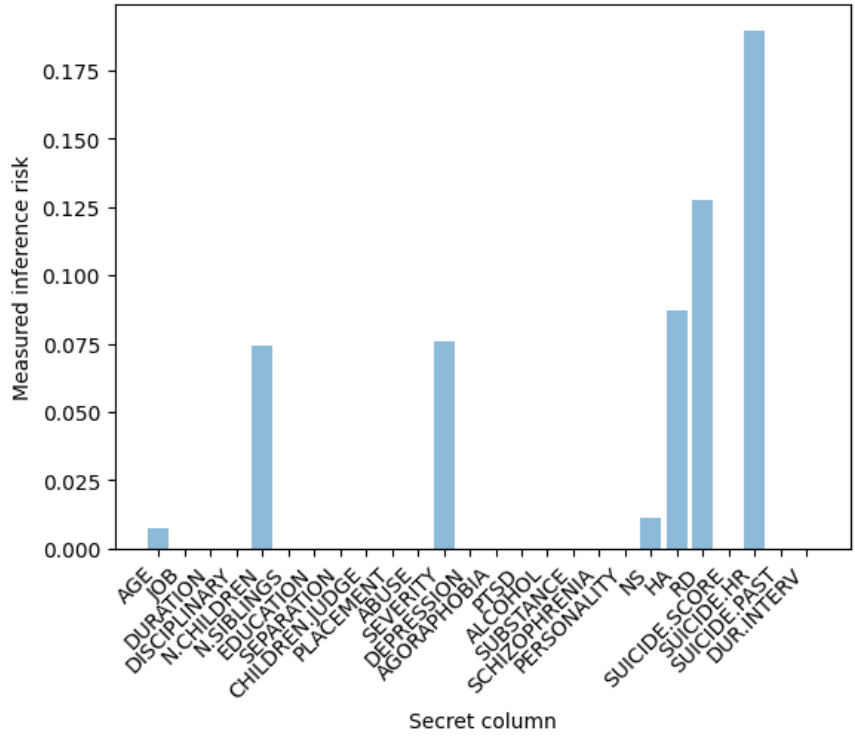

*Figure 16. Scores obtained by $S_2$ after simulated inference attacks with each variable being successively selected as a target.*

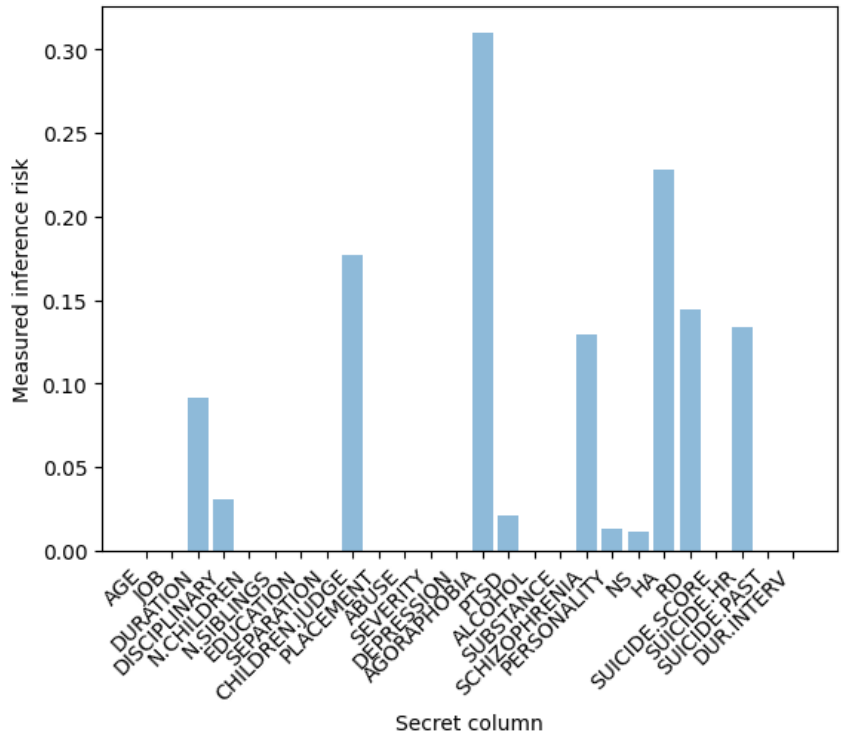

*Figure 17. Scores obtained by $S_3$ after simulated inference attacks with each variable being successively selected as a target.*

**Utility-related assessment**

Bivariate distributions of $D_O$ and of the three synthetic datasets for numerical variables are plotted in Figure 18. The univariate distributions of $S_1$, $S_2$ and $S_3$ plotted against $D_O$ are shown in Figures 19 to 21. Overall, all these distributions were respected by synthetic datasets, with the possible exception of EDUCATION and SEVERITY for which $S_2$ and $S_3$ presented slight differences with $D_O$. Interestingly, the filtering process tended to select units replicating an anomalous value for DUR.INTERV (i.e. a null duration), that escaped data management. This may be due to the large gain in Mahalanobis distance provided by such outlying values, which emphasizes the importance of meticulous management operations prior to data synthesis.

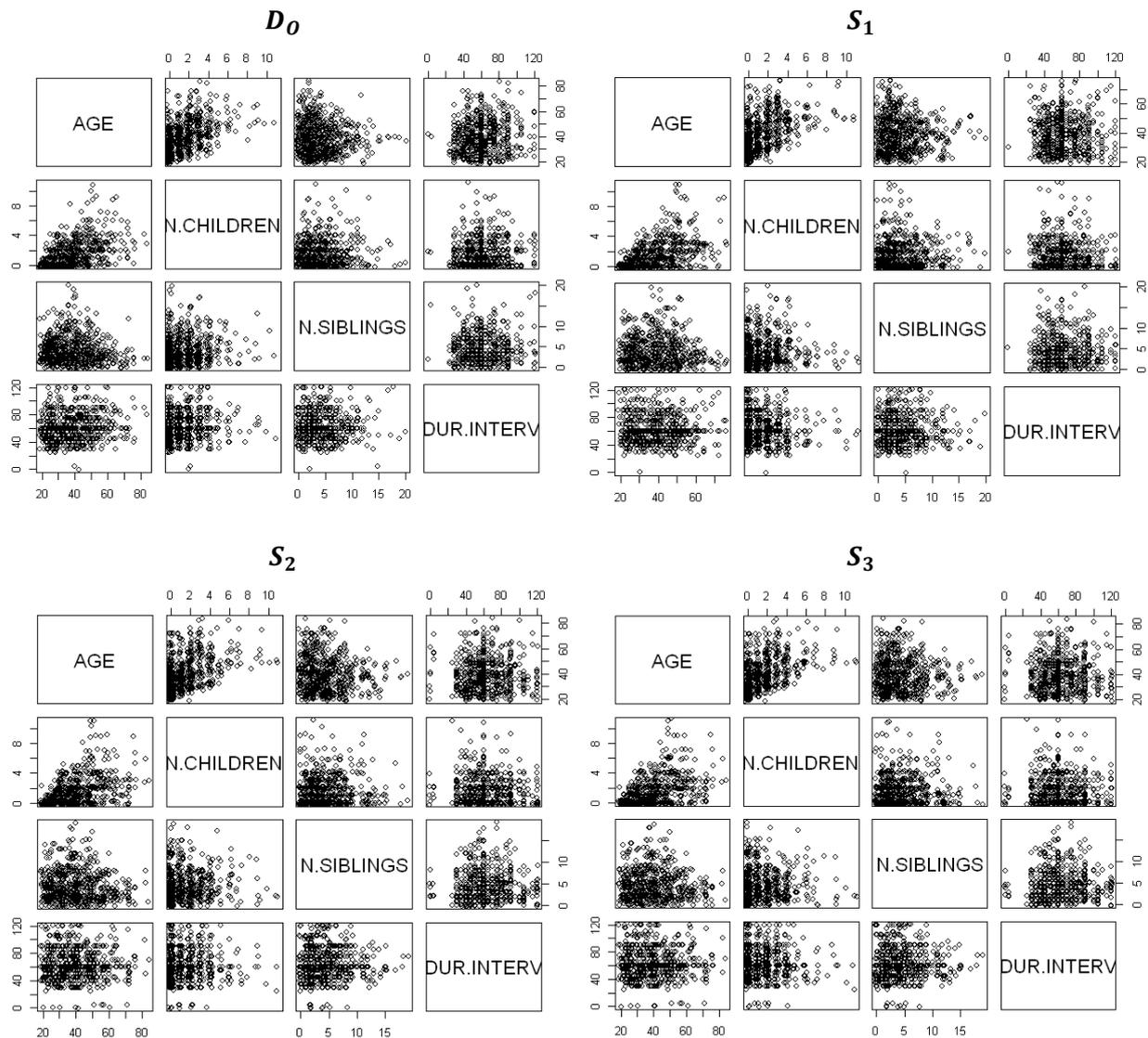

*Figure 18. Scatter plots of numerical variables for $D_o$ and each synthetic dataset.*

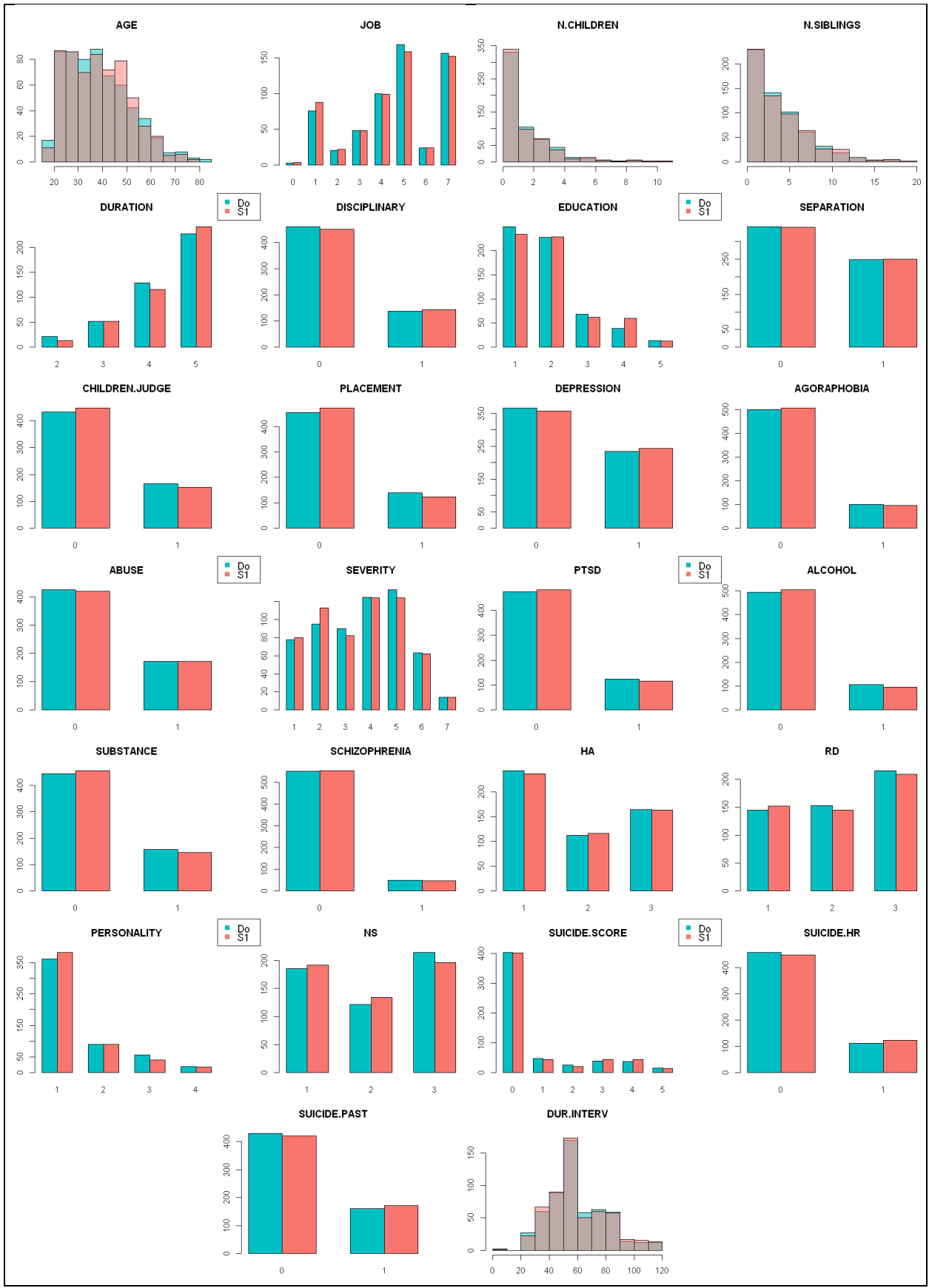

Figure 19. Univariate distributions of $S_1$ plotted against $D_o$.

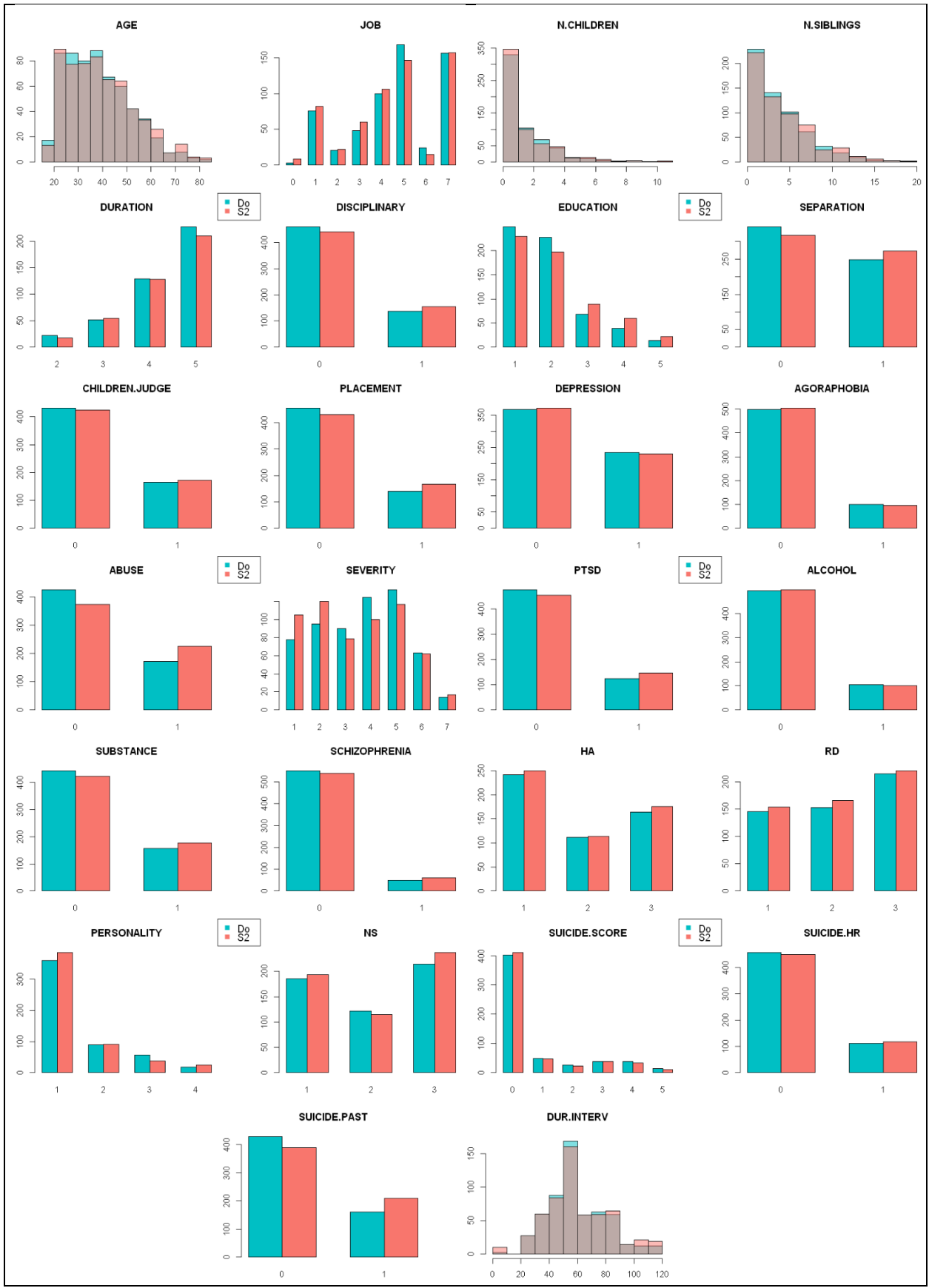

Figure 20. Univariate distributions of $S_2$ plotted against $D_o$.

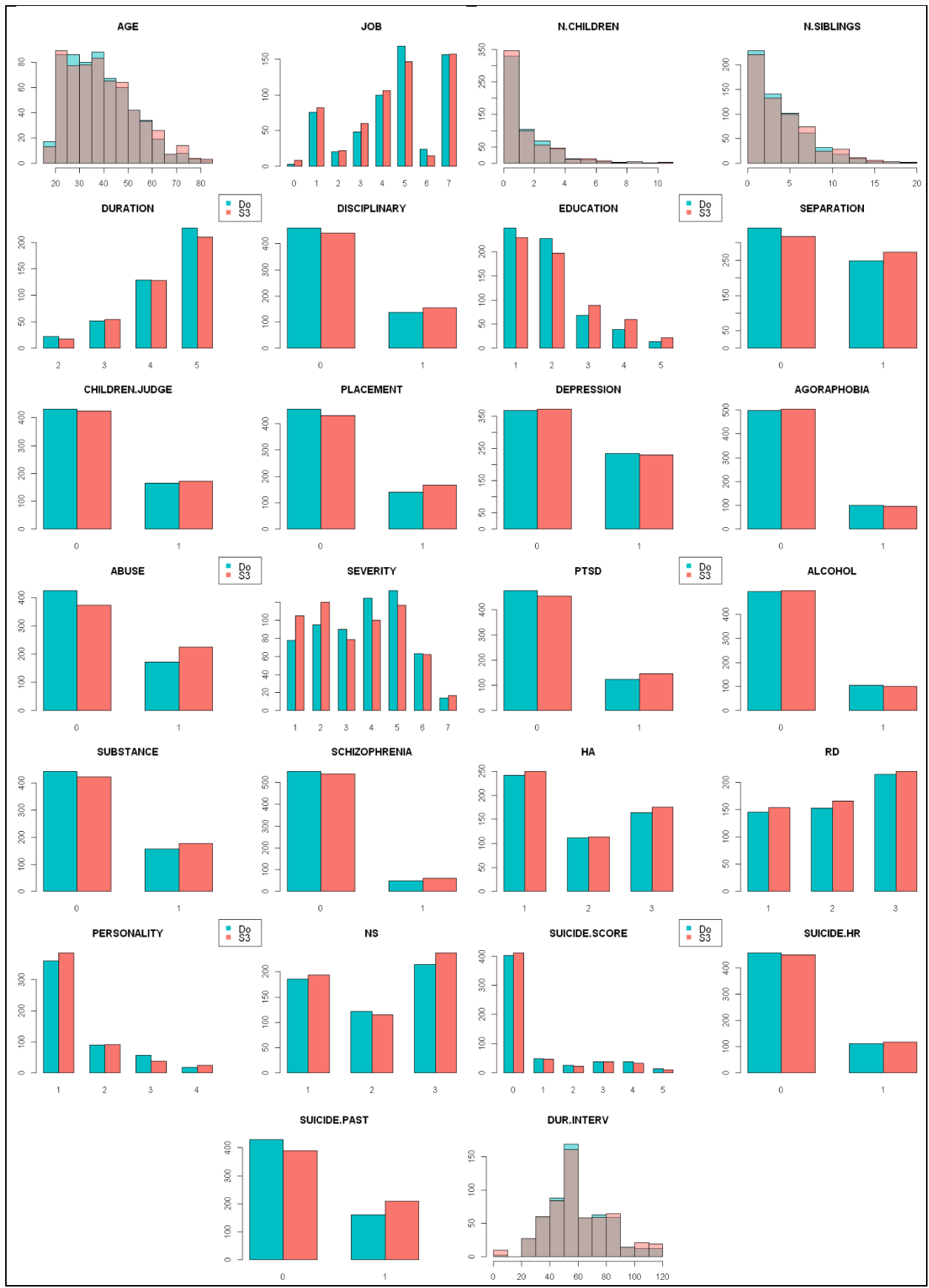

Figure 21. Univariate distributions of $S_3$ plotted against $D_o$.

Prevalence data for seven diagnoses of the DSM-IV were reconstituted following the Purdam and Elliot methodology. They are shown in Table 4 and Figure 22. All synthetic datasets performed equally well in this respect, providing estimates close to those of $D_O$, with the notable exception of agoraphobia and schizophrenia, for which every synthetic estimates were close to the half of the $D_O$ estimate. Interestingly, this was precisely for these diagnoses that sampling fluctuations caused $D_O$ to overestimate original prevalence rates.

|  | $D_I$ | $D_o$ | $S_1$ | $S_2$ | $S_3$ |
|---|---|---|---|---|---|
| Major depressive disorder | 0.18 | 0.23 | 0.23 | 0.20 | 0.20 |
| Agoraphobia | 0.07 | 0.10 | 0.05 | 0.05 | 0.05 |
| Post traumatic stress disorder | 0.10 | 0.14 | 0.14 | 0.15 | 0.15 |
| Alcohol / substance dependence | 0.14 | 0.15 | 0.13 | 0.15 | 0.15 |
| Alcohol dependence | 0.09 | 0.09 | 0.07 | 0.08 | 0.08 |
| Substance dependence | 0.11 | 0.12 | 0.09 | 0.11 | 0.11 |
| Schizophrenia | 0.04 | 0.07 | 0.03 | 0.02 | 0.02 |

*Table 4. Prevalence data reconstituted using the Purdam and Elliot methodology. These prevalence rates were calculated in inmates having a SEVERITY score of at least 5. $D_I$ refers to the complete data of 799 observations used by the authors of the original study.*

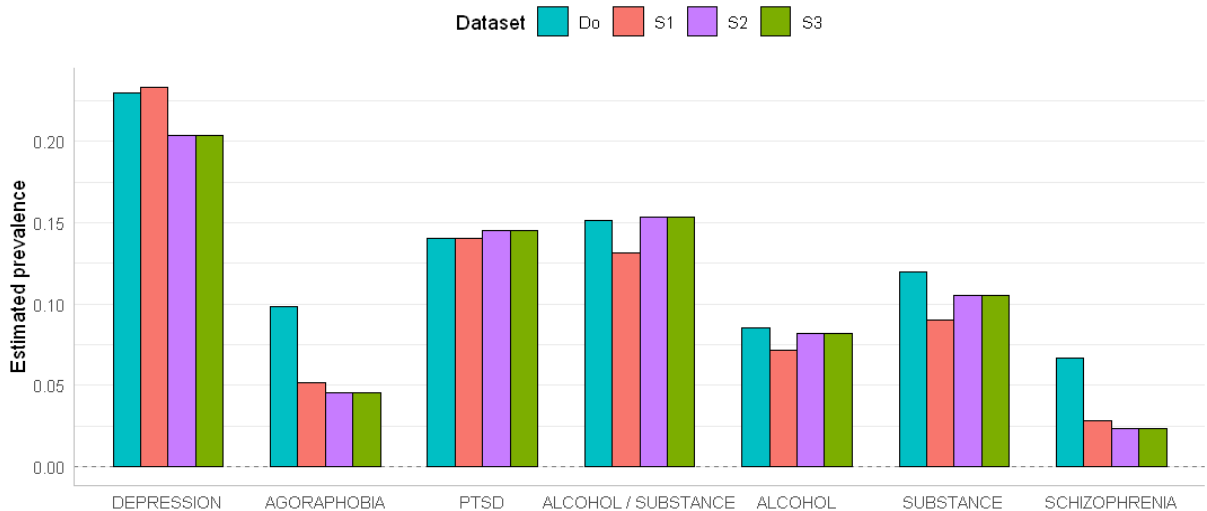

*Figure 22. Same data as in Table 4 plotted for $D_o$, $S_1$, $S_2$ and $S_3$.*

As planned in our assessment protocol, pMSE were determined for each synthetic dataset, resulting without surprise in $S_2$ (pMSE=0.123) and $S_3$ (pMSE=0.124) demonstrating a bit less utility than $S_1$ (pMSE=0.109). This finding is confirmed and illustrated by standardized pMSE ratios plotted in Figure 23, 24 and 25. More surprisingly, these ratios were especially high in $S_1$ for the pair AGORAPHOBIA × DEPRESSION, which may be related to the underestimate of the prevalence rate of agoraphobia when using this dataset.

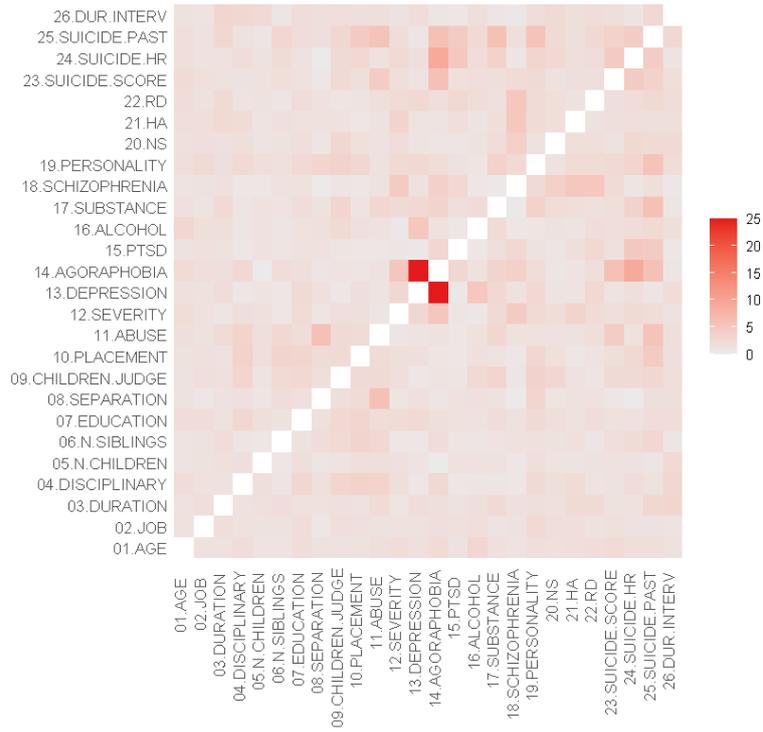

*Figure 23. Matrix of standardized pMSE ratios for each pair of variables in $S_1$.*

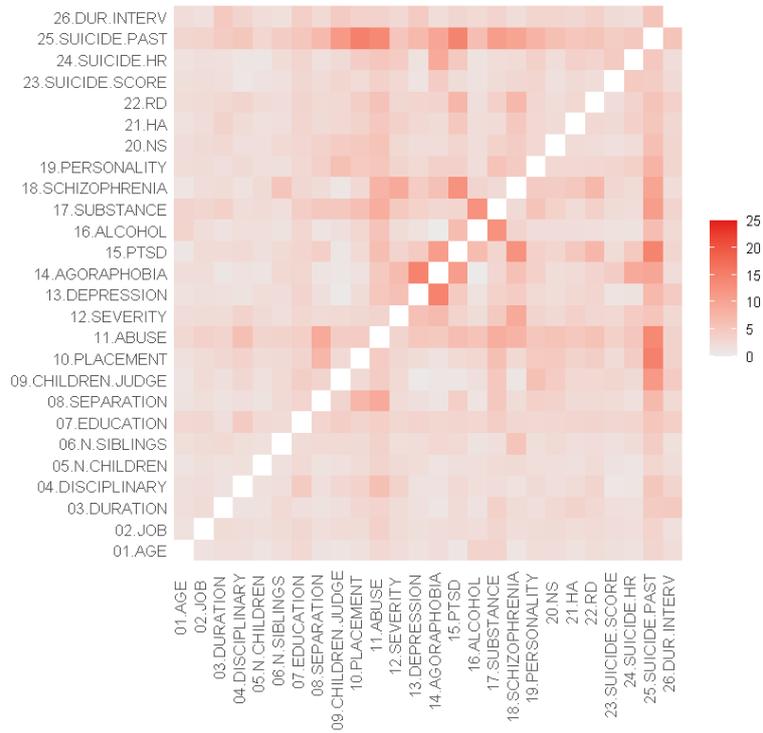

*Figure 24. Matrix of standardized pMSE ratios for each pair of variables in $S_2$.*

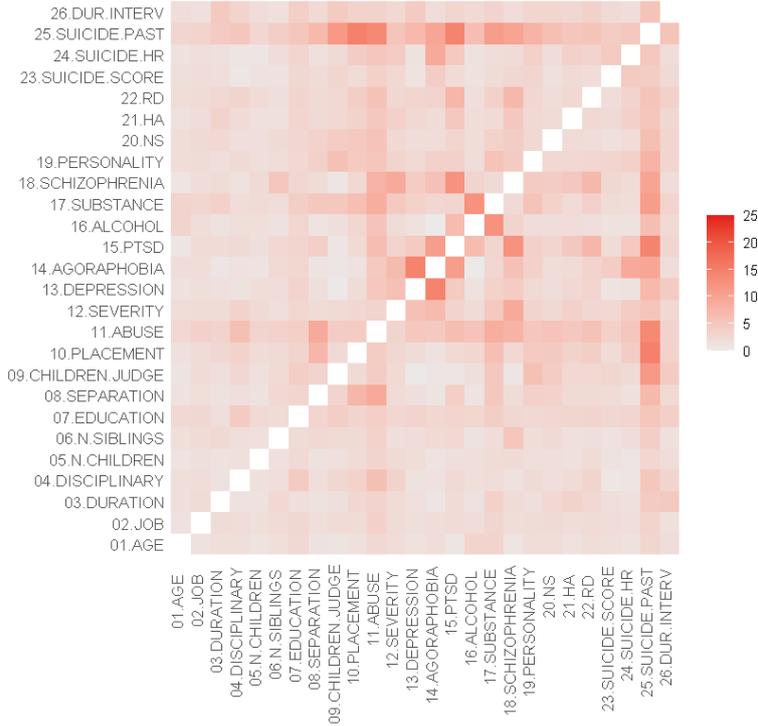

*Figure 25. Matrix of standardized pMSE ratios for each pair of variables in $S_3$.*

# Discussion

Throughout this work, we showed how a multi-step SDGF can produce high fidelity data while preserving the confidentiality of individuals from which they originate. Both these qualities have been assessed on a rich dataset featuring complex joint distributions. We used both proven and innovative assessment methods which gave unequivocal results. In particular, the calculated pMSE as well as the GTCAP were found to be persuasively low in all synthetic datasets generated. Dataset $S_2$ was close to $S_1$ for all metrics computed, and performed better on most privacy metrics, which demonstrates the relevance of our distance-based filtering method (aiming, among other, at preventing identification disclosure and related attacks). $S_3$ and $S_2$ had a nearly identical content, which shows that individuals of $D_O$ could be protected against MDA at a low utility cost.

These results are encouraging and show that the Open-CESP project is technically feasible, at least for some datasets. This, in our view, is a great step toward compliance with the core Open Science principles. However, research is still needed to determine what are the conditions of this technical feasibility. For most types of disclosure risks, our assessment method has indeed a weak formal grounding. Consequently, there is always an open possibility that another dataset could not have resulted in such neat outcomes. While waiting for the establishment of formal foundations to assess the intrinsic qualities of SDGF, replicating this work with other original datasets could give us precious insights on these validity conditions. In parallel, the properties of the novel metrics that we introduced, such as the GTCAP, should be established more thoroughly. This hopefully could help determining the best risk-utility tradeoff for upcoming synthetic datasets, which at the time being is still a major research challenge (83).

Another major contribution of this work is the introduction of a formal protection framework against MDA. This in our knowledge had never been reported before. By showing that high levels of protection against these attacks can be provided by a minimal noise addition, we sketched out a solution to what should be a legitimate concern of every data provider. The solution is not complete however, because this framework in its current state still has substantial limitations. First and foremost, its validity conditions need to be studied in more details, at least to answer the following questions: how to deal with values at the extreme periphery of the distribution? How to deal with small samples and/or small populations? How to deal with non-standard distributions? An important development would also be to treat the multivariate case, as a typical attacker would arguably use more than one variable to conduct a MDA. Other open questions include the feasibility of a noise addition framework preserving more of the variance of the original data, and the specification of an optimal algorithm to estimate ECAP and related values. In the meantime, comparing the current framework against existing membership disclosure metrics such as the F1 score (37) seems to be a necessary step to validate it further.

Given these results, and even if further developments are undoubtedly to come in the following years, we think that the Open-CESP initiative can be launched at a full-scale as of now. In this respect, legitimate concerns are to be anticipated. Despite their intuitive appeal, the probabilistic nature of the data synthesis and most of the additional steps of our SDGF may generate the fear that a privacy attack could be successful, even by chance. These concerns are theoretically well-founded (32). However, they need to be replaced in the larger context of medical research and data protection. When researchers communicate on their results, they release data, which can always lead to disclosure with a nonzero probability. Similarly, physical data storage locations can be prone to privacy breaches. However, research data are still communicated and stored worldwide. Thus there is always a trade-off made between the risk of disclosure and the expected benefits of using this type of data. Synthetic data make no exception, and as the conditions of this trade-off are more and more understood, we think that their large scale public release has nothing unrealistic.

## Conclusion

By successfully assessing the quality of data produced using a novel multi-step synthetic data generation framework, we showed the technical soundness of the Open-CESP project. In doing so, we also made several more general contributions including a novel formal protection framework against membership disclosure attacks as well as a generalized privacy metric compatible with various data types. This work opens the door to further investigation related to the determination of the best risk-utility tradeoff for synthetic data. Despite these persisting challenges, the Open-CESP initiative seems ripe for full-scale implementation, thus aligning with the broad Open Science objectives. Further work will refine the methods we used, but the path forward for public release of synthetic data appears promising.

# Table of abbreviations

| | |
|---|---|
| CART | Classification And Regression Trees |
| CESP | Centre de recherche en Epidémiologie et Santé des Populations |
| CHB | Chronic Hepatitis B |
| CNIL | Commission Nationale de l'Informatique et des Libertés |
| COVID | Coronavirus Disease |
| DCAP | Differential Correct Attribution Probability |
| DSM-IV | Diagnostic and Statistical Manual of mental disorders – 4$^{th}$ edition |
| ECAP | Elemental Correct Attribution Probability |
| GTCAP | Generalized Targeted Correct Attribution Probability |
| KDE | Kernel Density Estimation |
| MDA | Membership Disclosure Attack |
| MSD | Mean Successive Difference |
| NIH | National Institutes of Health |
| NIPALS | Non-linear Iterative Partial Least Sqaures |
| PC1 | 1$^{st}$ Principal Component |
| PC2 | 2$^{nd}$ Principal Component |
| pMSE | Propensity score Mean Squared Error |
| RU | Risk-Utility |
| SDC | Statistical Disclosure Control |
| SDGF | Synthetic Data Generation Framework |
| SNDS | Système National des Données de Santé |
| TCAP | Targeted Correct Attribution Probability |
| TCI | Temperament and Character Inventory |